\documentclass{article}

\usepackage{tabularx}
\usepackage{microtype}
\usepackage{graphicx}
\usepackage{subfigure}
\usepackage{booktabs} 

\usepackage{hyperref}

\usepackage{amsmath,amssymb,graphicx,color,booktabs,bm,relsize,enumitem,multirow,amsthm,epsfig,caption,mathtools,xcolor}
\usepackage{dsfont}
\usepackage[export]{adjustbox}
\newsavebox{\measurebox}

\newcommand{\reviewer}[3]{
	\expandafter\newcommand\csname #1\endcsname[1]{
		\textcolor{#3}{[#2: ##1]}
	}
}
\reviewer{kimin}{KM}{purple}
\reviewer{younggyo}{YG}{blue}
\reviewer{lili}{LC}{violet}

\usepackage[accepted]{icml2021}

\icmltitlerunning{State Entropy Maximization with Random Encoders for Efficient Exploration}

\begin{document}

\twocolumn[
\icmltitle{State Entropy Maximization with Random Encoders \\ for Efficient Exploration}



\icmlsetsymbol{equal}{*}

\vspace{-0.0225in}
\begin{icmlauthorlist}
\icmlauthor{Younggyo Seo}{equal,kaist}
\icmlauthor{Lili Chen}{equal,ucb}
\icmlauthor{Jinwoo Shin}{kaist}
\icmlauthor{Honglak Lee}{umich,lg}
\icmlauthor{Pieter Abbeel}{ucb}
\icmlauthor{Kimin Lee}{ucb}
\end{icmlauthorlist}
\vspace{-0.0225in}

\icmlaffiliation{kaist}{KAIST}
\icmlaffiliation{ucb}{UC Berkeley}
\icmlaffiliation{umich}{University of Michigan}
\icmlaffiliation{lg}{LG AI Research}

\icmlcorrespondingauthor{Kimin Lee}{kiminlee@berkeley.edu}

\icmlkeywords{Machine Learning, ICML}

\vskip 0.3in
]



\printAffiliationsAndNotice{\icmlEqualContribution} 

\begin{abstract}
Recent exploration methods have proven to be a recipe for improving sample-efficiency in deep reinforcement learning (RL). However, efficient exploration in high-dimensional observation spaces still remains a challenge. 
This paper presents Random Encoders for Efficient Exploration (RE3), an exploration method that utilizes state entropy as an intrinsic reward.
In order to estimate state entropy in environments with high-dimensional observations,
we utilize a $k$-nearest neighbor entropy estimator in the low-dimensional representation space of a convolutional encoder.
In particular, we find that the state entropy can be estimated in a stable and compute-efficient manner by utilizing a randomly initialized encoder, which is fixed throughout training.
Our experiments show that RE3 significantly improves the sample-efficiency of both model-free and model-based RL methods on locomotion and navigation tasks from DeepMind Control Suite and MiniGrid benchmarks.
We also show that RE3 allows learning diverse behaviors without
extrinsic rewards, effectively improving sample-efficiency in downstream tasks.
Source code and videos are available at \url{https://sites.google.com/view/re3-rl}.
\end{abstract}

\section{Introduction}
Exploration remains one of the main challenges of deep reinforcement learning (RL) in complex environments with high-dimensional observations.
Many prior approaches to incentivizing exploration introduce intrinsic rewards based on a measure of state novelty.
These include count-based visitation bonuses \cite{bellemare2016unifying, tang2017exploration, ostrovski2017count} and prediction errors \cite{stadie2015incentivizing, houthooft2016vime, pathak2017curiosity, burda2018exploration, pathak2019self, sekar2020planning}. 
By introducing such novelty-based intrinsic rewards,
these approaches encourage agents to visit diverse states, 
but leave unanswered the fundamental question of how to quantify effective exploration in a principled way.

To address this limitation, \citet{lee2019efficient} and \citet{hazan2019provably} proposed that exploration methods should encourage uniform (i.e., maximum entropy) coverage of the state space.
For practical state entropy estimation without learning density models, \citet{mutti2020policy} estimate state entropy by measuring distances between states and their $k$-nearest neighbors. To extend this approach to high-dimensional environments, recent works \citep{tao2020novelty, badia2020never, liu2021behavior} have proposed to utilize the $k$-nearest neighbor state entropy estimator in a low-dimensional latent representation space. 
The latent representations are learned by auxiliary tasks such as dynamics learning \citep{tao2020novelty}, inverse dynamics prediction \citep{badia2020never}, and contrastive learning \citep{liu2021behavior}. However, these methods still involve optimizing multiple objectives throughout RL training.
Given the added complexity (e.g., hyperparameter tuning), instability, and computational overhead of optimizing auxiliary losses, it is important to ask whether effective state entropy estimation is possible without introducing additional learning procedures.

\begin{figure} [t!] \centering
\hfill
\includegraphics[width=0.46\textwidth]{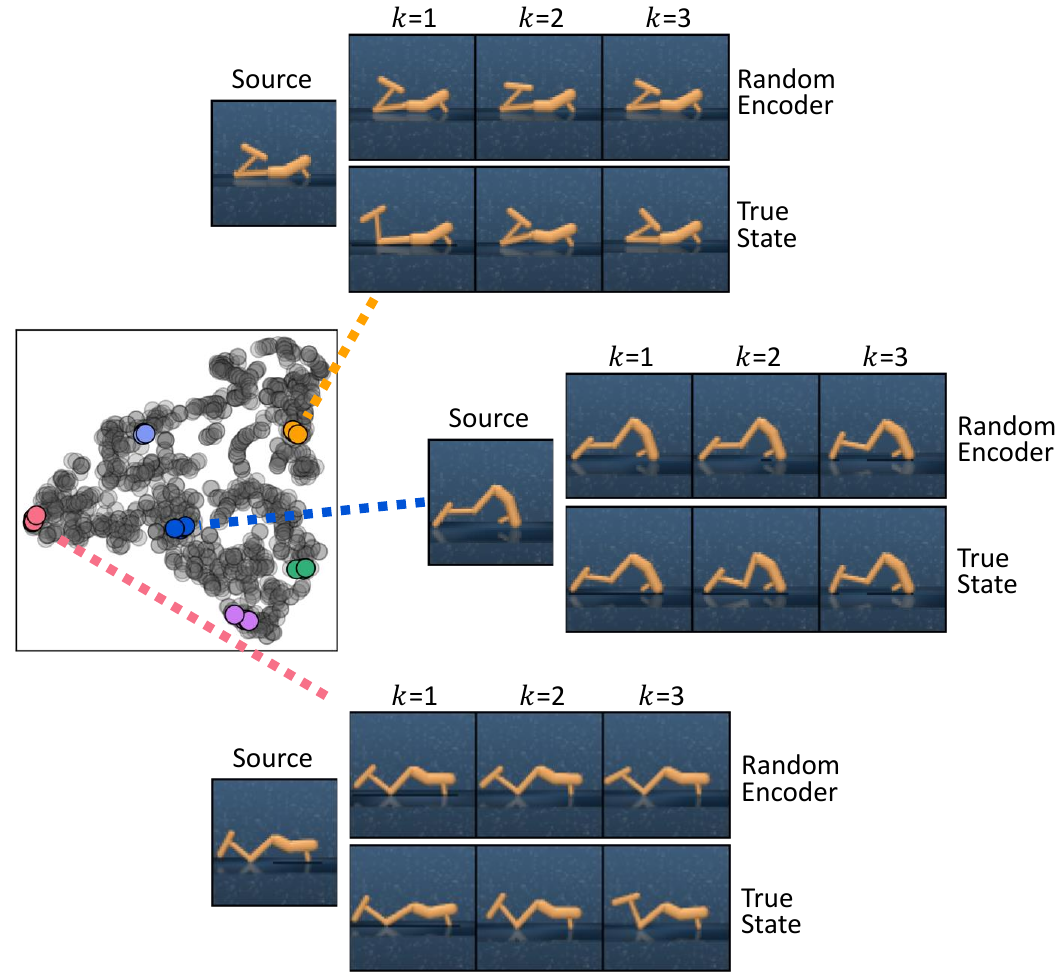}
\hfill
\vspace{-0.1in}
\caption{
Visualization of $k$-nearest neighbors of states found by measuring distances in the representation space of a randomly initialized encoder (Random Encoder) and ground-truth state space (True State) on the Hopper environment from DeepMind Control Suite \cite{tassa2020dm_control}.
We observe that the representation space of a random encoder effectively captures information about the similarity between states without any representation learning.
}
\label{fig:intro_random_encoder}
\vspace{-0.2in}
\end{figure}

In this paper, we present RE3: \textbf{R}andom \textbf{E}ncoders for \textbf{E}fficient \textbf{E}xploration, a simple, compute-efficient method for exploration without introducing additional models or representation learning.
The key idea of RE3 is to utilize a $k$-nearest neighbor state entropy estimator in the representation space of a randomly initialized encoder, which is fixed throughout training.
Our main hypothesis is that a randomly initialized encoder can provide a meaningful representation space for state entropy estimation by exploiting the strong prior of convolutional architectures. \citet{ulyanov2018deep} and \citet{caron2018deep} found that the structure alone of deep convolutional networks is a powerful inductive bias that allows relevant features to be extracted for tasks such as image generation and classification. 
In our case, we find that the representation space of a randomly initialized encoder effectively captures information about similarity between states, as shown in Figure~\ref{fig:intro_random_encoder}.
Based upon this observation, we propose to maximize a state entropy estimate in the fixed representation space of a randomly initialized encoder.

\newpage

We highlight the main contributions of this paper below:
\begin{itemize}[topsep=1.0pt,itemsep=0.85pt]
    \item [$\bullet$] RE3 significantly improves the sample-efficiency of both model-free and model-based RL methods
    on widely used DeepMind Control Suite \cite{tassa2020dm_control}, MiniGrid \cite{gym_minigrid}, and Atari \cite{bellemare2013arcade} benchmarks.
    \item [$\bullet$] RE3 encourages exploration without introducing representation learning or additional models, outperforming state entropy maximization schemes that involve representation learning and exploration methods that introduce additional models for exploration \cite{pathak2017curiosity, burda2018exploration}.
    \item [$\bullet$] RE3 is compute-efficient because it does not require gradient computations and updates for additional representation learning, making it a scalable and practical approach to exploration.
    \item [$\bullet$] RE3 allows learning diverse behaviors in environments without extrinsic rewards; 
    we further improve sample-efficiency in downstream tasks by fine-tuning a policy pre-trained with the RE3 objective.
\end{itemize}

\section{Related Work}
\label{sec:related_work}
{\bf Exploration in reinforcement learning. }
Exploration algorithms encourage the RL agent to visit a wide range of states by injecting noise to the action space~\cite{lillicrap2015continuous} or parameter space~\cite{fortunato2017noisy, plappert2017parameter}, maximizing the entropy of the action space~\cite{ziebart2010modeling, haarnoja2018soft}, and setting diverse goals that guide exploration~\cite{florensa2018automatic, nair2018visual, pong2019skew, colas2019curious}.
Another line of exploration algorithms introduce intrinsic rewards proportional to prediction errors~\cite{houthooft2016vime,pathak2017curiosity, burda2018exploration, sekar2020planning}, and count-based state novelty~\cite{bellemare2016unifying, tang2017exploration, ostrovski2017count}.
Our approach differs in that we explicitly encourage the agent to uniformly visit all states by maximizing the entropy of the state distribution, instead of depending on metrics from additional models.

{\bf State entropy maximization. }
Most closely related to our work are methods that maximize the entropy of state distributions.
\citet{hazan2019provably,lee2019efficient} proposed to maximize state entropy estimated by approximating the state density distribution.
Instead of approximating complex distributions, 
\citet{mutti2020policy} proposed to maximize a $k$-nearest neighbor state entropy estimate from on-policy transitions.
Recent works extend this method to environments with high-dimensional observations.
\citet{tao2020novelty} employ model-based RL techniques to build a representation space for the state entropy estimate that measures similarity in dynamics,
and \citet{badia2020never} proposed to measure similarity in the representation space learned by inverse dynamics prediction.
The work closest to ours is \citet{liu2021behavior}, which uses off-policy RL algorithms to maximize
the $k$-nearest neighbor state entropy estimate in contrastive representation space~\cite{srinivas2020curl}
for unsupervised pre-training.
We instead explore the idea of utilizing a fixed random encoder to obtain a stable entropy estimate without any representation learning.

{\bf Random encoders. }
Random weights have been utilized in neural networks since their beginnings, most notably in a randomly initialized first layer \citep{gamba1961further} termed the Gamba perceptron by \citet{minsky1969perceptrons}.
Moreover, nice properties of random projections are commonly exploited for low-rank approximation \citep{vempala2005random, rahimi2007random}.
These ideas have since been extended to deep convolutional networks, where random weights are surprisingly effective at image generation and restoration \citep{ulyanov2018deep}, image classification and detection \cite{caron2018deep}, and fast architecture search \citep{saxe2011random}. 
In natural language processing, \citet{wieting2019no} demonstrated that learned sentence embeddings show marginal performance gain over random embeddings.
In the context of RL, \citet{gaier2019weight} showed that competitive performance can be achieved by architecture search over random weights without updating weights, and \citet{lee2019network} utilized randomized convolutional neural networks to improve the generalization of deep RL agents.
Building on these works, we show that random encoders can also be useful for efficient exploration in environments with high-dimensional observations.

\begin{figure*} [t!] \centering
\includegraphics[width=0.775\textwidth]{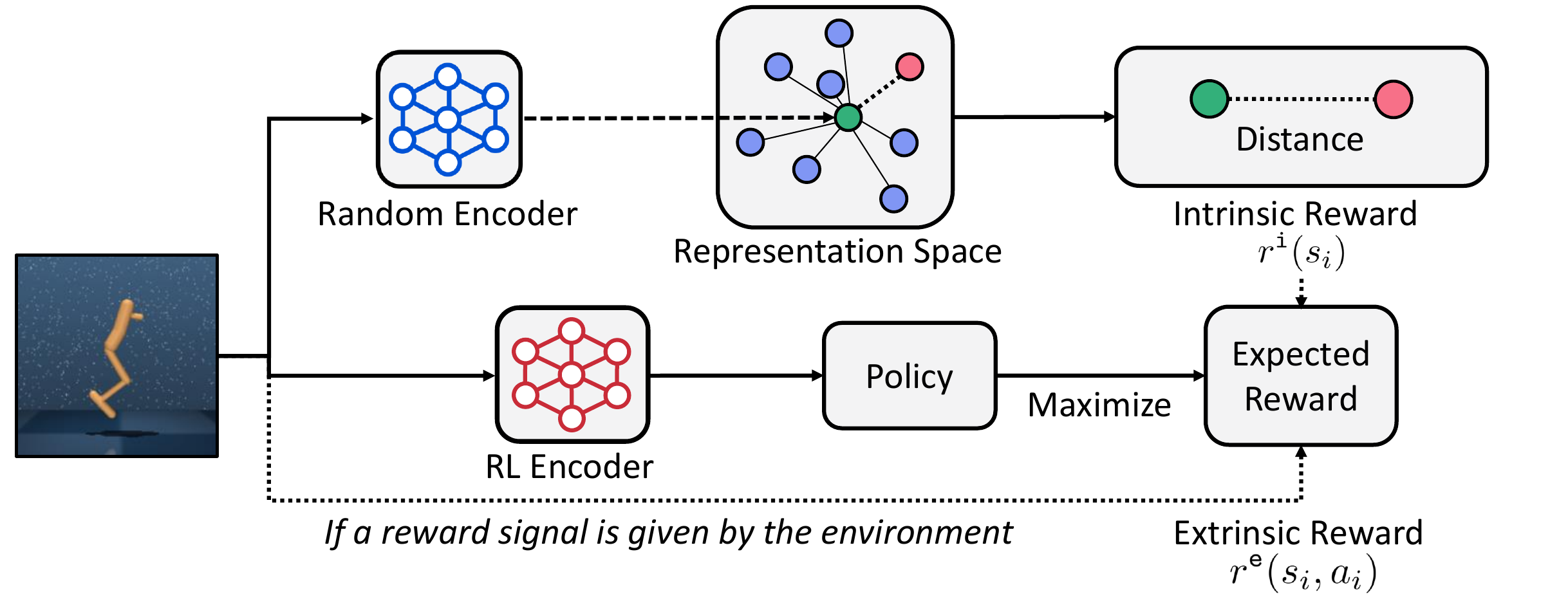}
\vspace{-0.125in}
\caption{Illustration of our approach. The intrinsic reward for each observation is computed as the distance to its $k$-nearest neighbor, measured between low-dimensional representations obtained from the fixed random encoder. The intrinsic reward is then combined with extrinsic reward from the environment, if present. A separate RL encoder is introduced for a policy that maximizes expected reward.}
\vspace{-0.1in}
\label{fig:method_overview}
\end{figure*}

\section{Method}
\subsection{Preliminaries}
We formulate a control task with high-dimensional observations as a partially observable Markov decision process (POMDP; \citealt{sutton2018reinforcement, kaelbling1998planning}), which is defined as a tuple $\left( \mathcal{O}, \mathcal{A}, p, r^{\tt{e}}, \gamma\right)$.
Here, $\mathcal{O}$ is the high-dimensional observation space,
$\mathcal{A}$ is the action space,
$p\left(o^\prime|o_{\leq t}, a_{t}\right)$ is the transition dynamics,
$r^{\tt{e}}: \mathcal{O} \times \mathcal{A} \rightarrow \mathbb{R}$ is the reward function that maps the current observation and action to a reward $r^{\tt{e}}_{t} = r^{\tt{e}}\left(o_{\leq t}, a_{t}\right)$,
and $\gamma \in [0,1)$ is the discount factor.
By following common practice \cite{mnih2015human}, we reformulate the POMDP as an MDP \cite{sutton2018reinforcement} by stacking consecutive observations into a state $s_{t} = \{o_{t}, o_{t-1}, o_{t-2}, ...\}$. 
For simplicity of notation, we redefine the reward function as $r^{\tt{e}}_{t} = r^{\tt{e}}\left(s_{t},a_{t}\right)$.
The goal of RL is to learn a policy $\pi(a_{t}|s_{t})$ that maximizes the expected return defined as the total accumulated reward.

{\bf $\bm{k}$-nearest neighbor entropy estimator. }
Let $X$ be a random variable with a probability density function $p$ whose support is a set $\mathcal{X} \subset \mathbb{R}^{q}$. 
Then its differential entropy is given as $\mathcal{H}(X) = -\mathbb{E}_{x \sim p(x)}[\log p(x)]$.
When the distribution $p$ is not available, this quantity can be estimated given $N$ i.i.d realizations of $\{x_{i}\}^{N}_{i=1}$ \cite{beirlant1997nonparametric}.
However, since it is difficult to estimate $p$ with high-dimensional data, particle-based $k$-nearest neighbors ($k$-NN) entropy estimator \cite{singh2003nearest} can be employed:
\begin{align}
    \widehat{\mathcal{H}}^{k}_{N}(X) &= \frac{1}{N} \sum^{N}_{i=1} \log \frac{N \cdot ||x_{i} - x_{i}^{\text{$k$-NN}}||_{2}^{q} \cdot \widehat{\pi}^{\frac{q}{2}}}{k \cdot \Gamma(\frac{q}{2} + 1)} + C_{k}
    \label{eq:knn_state_entropy_estimator}\\
    &\propto \frac{1}{N}\sum^{N}_{i=1} \log ||x_{i} - x_{i}^{\text{$k$-NN}}||_{2}, 
    \label{eq:simplied_knn_state_entropy_estimator}
 \end{align}
where $x_{i}^{\text{$k$-NN}}$ is the $k$-NN of $x_{i}$ within a set $\{x_{i}\}_{i=1}^{N}$, $C_{k} = \log k - \Psi(k)$ a bias correction term, $\Psi$ the digamma function, $\Gamma$ the gamma function, $q$ the dimension of $x$, $\widehat{\pi} \approx 3.14159$, and the transition from (\ref{eq:knn_state_entropy_estimator}) to (\ref{eq:simplied_knn_state_entropy_estimator}) always holds for $q > 0$.

\subsection{Random Encoders for Efficient Exploration}
We present Random Encoders for Efficient Exploration (RE3), which encourages exploration in high-dimensional observation spaces by maximizing state entropy.
The key idea of RE3 is $k$-nearest neighbor entropy estimation in the low-dimensional representation space of a randomly initialized encoder.
To this end, we propose to compute the distance between states in the representation space of a random encoder $f_{\theta}$ whose parameters $\theta$ are randomly initialized and fixed throughout training.
The main motivation arises from our observation that distances in the representation space of $f_{\theta}$ are already useful for finding similar states without any representation learning (see Figure~\ref{fig:intro_random_encoder}).

{\bf State entropy estimate as intrinsic reward. }
To define the intrinsic reward proportional to state entropy estimate by utilizing (\ref{eq:simplied_knn_state_entropy_estimator}), we follow the idea of \citet{liu2021behavior} that treats each transition as a particle, hence our intrinsic reward is given as follows:
\begin{align}
    r^{\tt{i}}(s_{i}) := \log(||y_{i} - y^{\text{$k$-NN}}_{i}||_{2} + 1),
    \label{eqn:re3_intrinsic}
\end{align}
where $y_{i} = f_{\theta}(s_{i})$ is a fixed representation from a random encoder 
and $y_{i}^{\text{$k$-NN}}$ is the $k$-nearest neighbor of $y_{i}$ within a set of $N$ representations $\{y_{1}, y_{2}, ..., y_{N}\}$.
Here, our intuition is that measuring the distance between states in the fixed representation space produces a more stable intrinsic reward as the distance between a given pair of states does not change during training. To compute distances in latent space in a compute-efficient manner, we propose to additionally store low-dimensional representations $y$ in the replay buffer $\mathcal{B}$ during environment interactions. Therefore, we avoid processing high-dimensional states through an encoder for obtaining representations at every RL update. Moreover, we can feasibly compute the distance of $y_{i}$ to all entries $y \in \mathcal{B}$, in contrast to existing approaches that utilize on-policy samples \cite{mutti2020policy}, or samples from a minibatch \cite{liu2021behavior}.
Our scheme enables stable, precise entropy estimation in a compute-efficient manner.

\begin{figure*} [t!] \centering
\hfill
\subfigure[Walker]
{
\includegraphics[width=0.1425\textwidth]{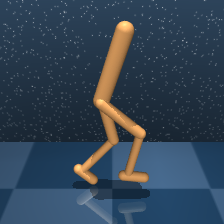}
\label{fig:dm_control_example_walker}} 
\hfill
\subfigure[Hopper]
{
\includegraphics[width=0.1425\textwidth]{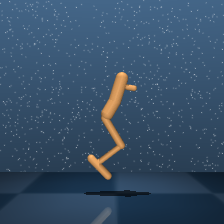}
\label{fig:dm_control_example_hopper}} 
\hfill
\subfigure[Quadruped]
{
\includegraphics[width=0.1425\textwidth]{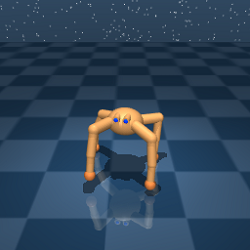}
\label{fig:dm_control_example_quadruped}} 
\hfill
\subfigure[Cheetah]
{
\includegraphics[width=0.1425\textwidth]{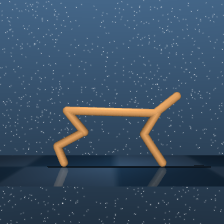}
\label{fig:dm_control_example_cheetah}} 
\hfill
\subfigure[Cartpole]
{
\includegraphics[width=0.1425\textwidth]{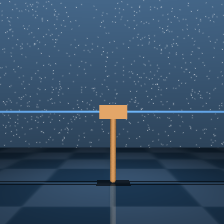}
\label{fig:dm_control_example_cartpole}} 
\hfill
\subfigure[Pendulum]
{
\includegraphics[width=0.1425\textwidth]{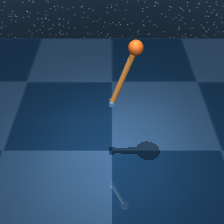}
\label{fig:dm_control_example_pendulum}} 
\hfill
\\
\vspace{-0.15in}
\caption{Image observations for visual control tasks from DeepMind Control Suite
\cite{tassa2020dm_control} used in our experiments. 
The high-dimensionality of these observations necessitates an efficient method for state entropy estimation.}
\label{fig:dm_control_example}
\vspace{-0.15in}
\end{figure*}

{\bf The RE3 objective. }
We propose to utilize the intrinsic reward $r^{\tt{i}}$ 
for (a) \textit{online RL}, where the agent solves target tasks guided by extrinsic reward $r^{\tt{e}}$ from environments, 
and (b) \textit{unsupervised pre-training}, where the agent learns to explore the high-dimensional observation space in the absence of extrinsic rewards, i.e., $r^{\tt{e}} = 0$. This exploratory policy from pre-training, in turn, can be used to improve the sample-efficiency in downstream tasks by fine-tuning.
Formally, we introduce a policy $\pi_{\phi}$, parameterized by $\phi$, that maximizes the expected return $\mathbb{E}_{\pi_{\phi}}\left[ \sum^{\infty}_{j=0} \gamma^{j} r^{\tt{total}}_{j} \right]$, where the total reward $r^{\tt{total}}_{j}$ is defined as:
\begin{align}
    r^{\tt{total}}_{j} := r^{\tt{e}} (s_{j}, a_{j}) + \beta_{t} \cdot r^{\tt{i}} (s_{j}), 
    \label{eqn:combined_reward}
\end{align}
where $\beta_{t}\geq 0$ is a hyperparameter that determines the tradeoff between exploration and exploitation at training timestep $t$.  We use the exponential decay schedule for $\beta_{t}$ throughout training to encourage the agent to further focus on extrinsic reward from environments as training proceeds, i.e., $\beta_{t} = \beta_{0}(1-\rho)^t$, where $\rho$ is a decay rate.
While the proposed intrinsic reward would converge to 0 as more similar states are collected during training, we discover that decaying $\beta_{t}$ empirically stabilizes the performance. We provide the full procedure for RE3 with off-policy RL in Algorithm~\ref{alg:training} and on-policy RL in Algorithm~\ref{alg:on_policy}.

\begin{figure*} [t!] \centering
\includegraphics[width=0.96\textwidth]{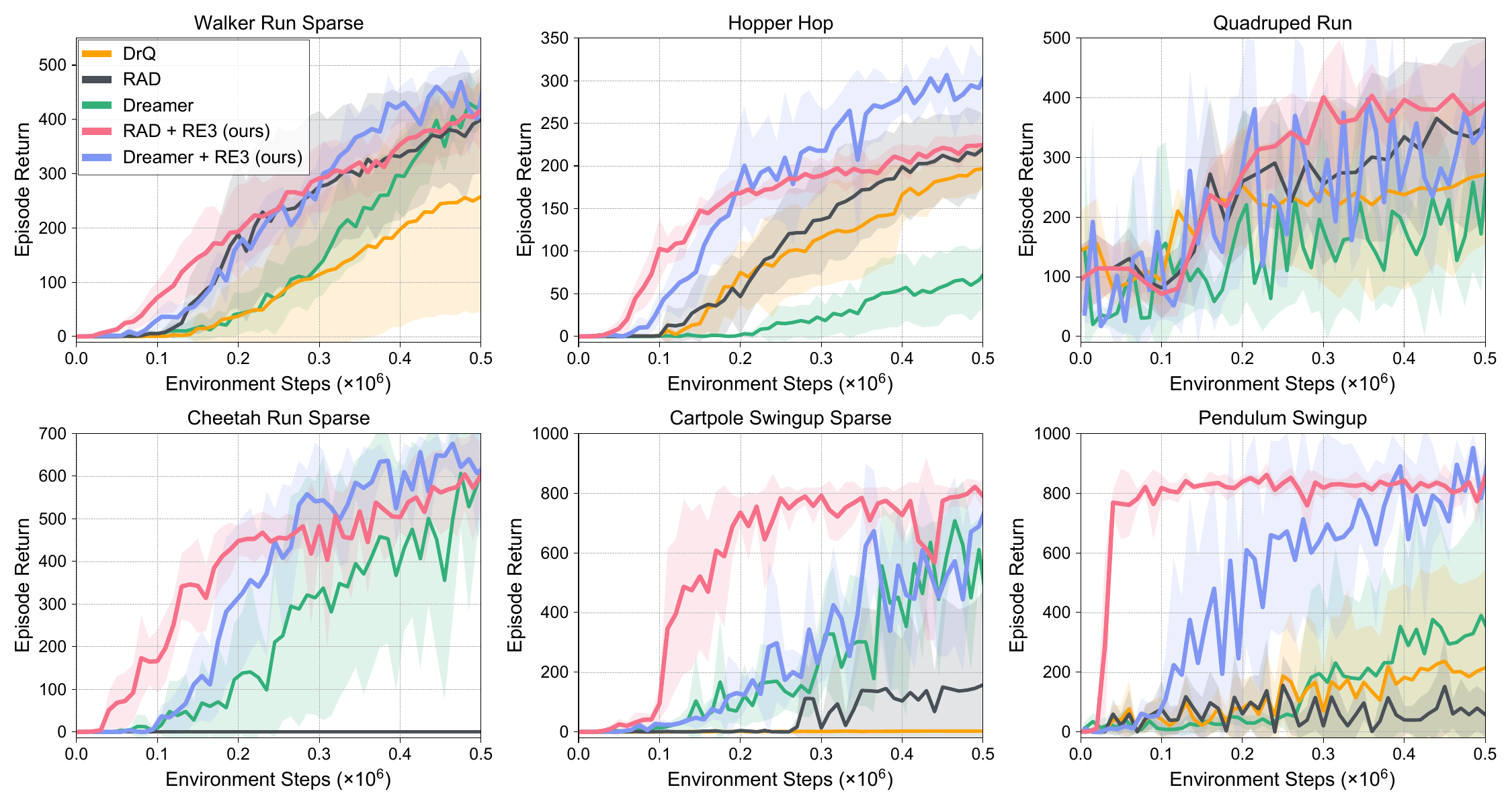}
\vspace{-0.18in}
\caption{Performance on locomotion tasks from DeepMind Control Suite. RE3 consistently improves the sample-efficiency of RAD and Dreamer. The solid line and shaded regions represent the mean and standard deviation, respectively, across five runs.}
\label{fig:dm_control_online}
\vspace{-0.1in}
\end{figure*}

\begin{figure*} [t!] \centering
\includegraphics[width=0.96\textwidth]{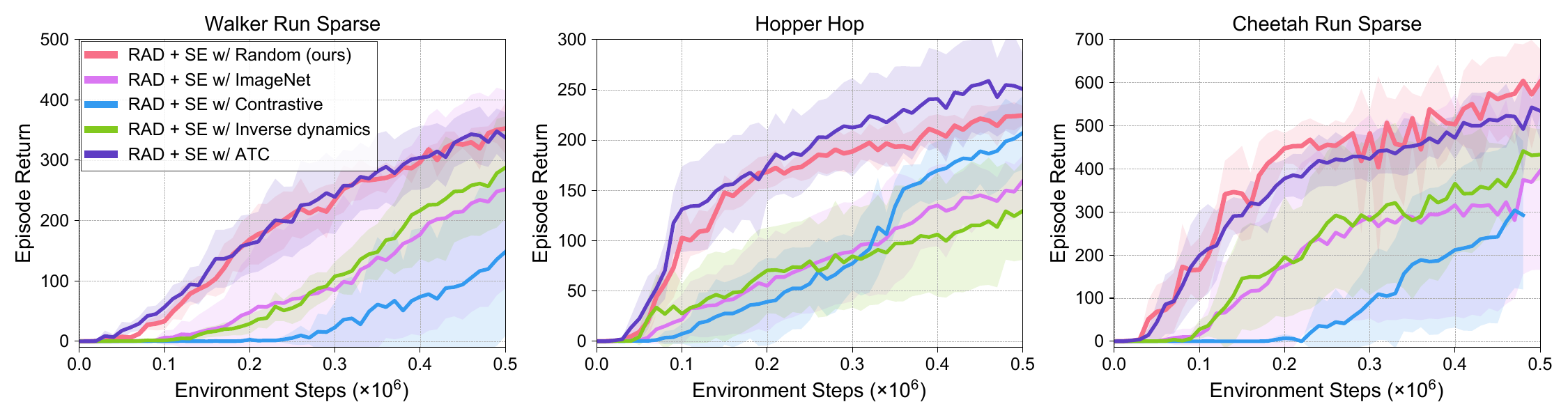}
\vspace{-0.18in}
\caption{We compare state entropy (SE) maximization with RE3 to state entropy maximization schemes that involve representation learning. The solid line and shaded regions represent the mean and standard deviation, respectively, across five runs.}
\label{fig:dm_control_representation}
\vspace{-0.15in}
\end{figure*}

\section{Experiments}
We designed experiments to answer the following questions:
\begin{itemize}[topsep=0.5pt]
    \item [$\bullet$] Can RE3 improve the sample-efficiency of both model-free and model-based RL algorithms (see Figure~\ref{fig:dm_control_online})?
    \item [$\bullet$] How does RE3 compare to state entropy maximization schemes that involve representation learning (see Figure~\ref{fig:dm_control_representation}) and other exploration schemes that introduce additional models for exploration (see Figure~\ref{fig:dm_control_exploration})?
    \item [$\bullet$] How compute-efficient is RE3 (see Figure~\ref{fig:dm_control_analysis_compute_efficiency})?
    \item [$\bullet$] Can RE3 further improve the sample-efficiency of off-policy RL algorithms by unsupervised pre-training (see Figure~\ref{fig:dm_control_behavior_visualization} and  Figure~\ref{fig:dm_control_unsupervised})?
    \item [$\bullet$] Can RE3 also improve the sample-efficiency of on-policy RL and off-policy RL in discrete control tasks (see Figure~\ref{fig:minigrid_all} and Figure~\ref{fig:atari_results})?
\end{itemize}

\subsection{DeepMind Control Suite Experiments}
{\bf Setup. }
To evaluate the sample-efficiency of our method, we compare to Dreamer \cite{hafner2019dream}, a state-of-the-art model-based RL method for visual control; and two state-of-the-art model-free RL methods, RAD \cite{laskin2020reinforcement} and DrQ \cite{kostrikov2020image}.
For comparison with other exploration methods, we consider RND \cite{burda2018exploration} and ICM \cite{pathak2017curiosity} that introduce additional models for exploration.
For RE3 and baseline exploration methods, we use RAD as the underlying model-free RL algorithm.
To further demonstrate the applicability of RE3 to model-based RL algorithms, we also consider a combination of Dreamer and RE3.
For random encoders, we use convolutional neural networks with the same architecture as underlying RL algorithms, but with randomly initialized parameters fixed during the training.
As for the newly introduced hyperparameters, we use $k = 3$, $\beta_{0} \in \{0.05, 0.25\}$, and $\rho \in \{0.0, 0.00001, 0.000025\}$.
We provide more details in Appendix~\ref{appendix:dm_control_detail}. Source code is available at \url{https://sites.google.com/view/re3-rl}.

\begin{algorithm}[t]
\caption{RE3: Off-policy RL version} \label{alg:training}
\begin{algorithmic}[1]
\STATE Initialize parameters of random encoder $\theta$, policy $\phi$
\STATE Initialize replay buffer $\mathcal{B} \leftarrow \emptyset$
\FOR{each timestep $t$}
\STATE {{\textsc{// Collect transitions}}}
\STATE Collect a transition $\tau_{t} = (s_{t}, a_{t}, s_{t+1}, r^{\tt{e}}_{t})$ from the interaction with the environment using policy $\pi_{\phi}$
\STATE Get a fixed representation $y_{t} = f_{\theta}(s_{t})$
\STATE $\mathcal{B} \leftarrow \mathcal{B} \cup \{(\tau_{t}, y_{t})\}$
\STATE {{\textsc{// Compute intrinsic reward}}}
\STATE Sample random minibatch $\{(\tau_{j}, y_{j})\}^{B}_{j=1} \sim \mathcal{B}$
\FOR{$j=1$ {\bfseries to} $B$}
\STATE Compute the distance $||y_{j} - y||_{2}$ for all representations $y \in \mathcal{B}$ and find the $k$-nearest neighbor $y_{j}^{\text{$k$-NN}}$
\STATE Compute $r_{j}^{\tt{i}} \leftarrow \log(||y_{j} - y_{j}^{\text{$k$-NN}}||_{2} + 1)$
\STATE Update $\beta_{t} \leftarrow \beta_{0}(1-\rho)^{t}$
\STATE Let $r^{\tt{total}}_{j} \leftarrow r_{j}^{\tt{e}} + \beta_{t} \cdot r_{j}^{\tt{i}}$
\ENDFOR
\STATE {{\textsc{// Update policy}}}
\STATE Update $\phi$ with transitions $\{(s_{j}, a_{j}, s_{j+1}, r_{j}^{\tt{total}})\}_{j=1}^{B}$
\ENDFOR
\end{algorithmic}
\end{algorithm}

\begin{figure*} [t!] \centering
\includegraphics[width=0.98\textwidth]{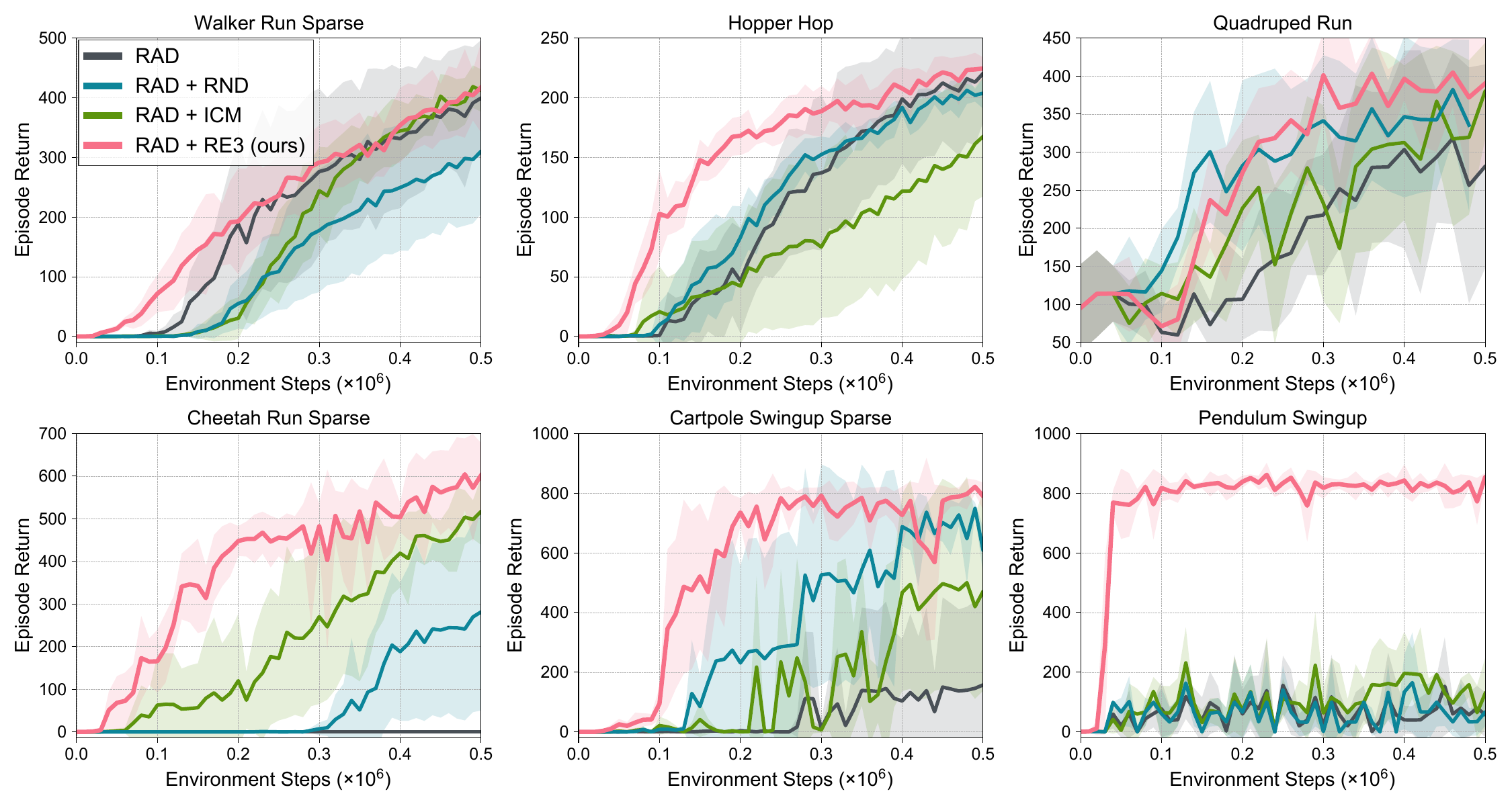}
\vspace{-0.18in}
\caption{Performance on locomotion tasks from DeepMind Control Suite. RAD + RE3 outperforms other exploration methods in terms of sample-efficiency. The solid line and shaded regions represent the mean and standard deviation, respectively, across five runs.}
\label{fig:dm_control_exploration}
\vspace{-0.1in}
\end{figure*}

{\bf Comparative evaluation. }
Figure~\ref{fig:dm_control_online} shows that RE3 consistently improves the sample-efficiency of RAD on various tasks.
In particular, RAD + RE3 achieves average episode return of 601.6 on Cheetah Run Sparse, where both model-free RL methods RAD and DrQ fail to solve the task.
We emphasize that state entropy maximization with RE3 achieves such sample-efficiency with minimal cost due to its simplicity and compute-efficiency.
We also observe that Dreamer + RE3 improves the sample-efficiency of Dreamer on most tasks, which demonstrates the applicability of RE3 to both model-free and model-based RL algorithms.

{\bf Effects of representation learning. }
To better grasp how RE3 improves sample-efficiency,
we compare to state entropy maximization schemes that involve representation learning in Figure~\ref{fig:dm_control_representation}.
Specifically, we consider a convolutional encoder trained by contrastive learning (RAD + SE w/ Contrastive), inverse dynamics prediction (RAD + SE w/ Inverse dynamics), and a ResNet-50 \cite{he2016deep} encoder pre-trained on ImageNet dataset (RAD + SE w/ ImageNet).
We found that our method (RAD + SE w/ Random) exhibits better sample-efficiency than approaches that continually update representations throughout training (RAD + SE w/ Contrastive, RAD + SE w/ Inverse dynamics).
This demonstrates that utilizing fixed representations helps improve sample-efficiency by enabling stable state entropy estimation throughout training.
We also observe that our approach outperforms RAD + SE w/ ImageNet,
implying that it is not necessarily beneficial to employ a pre-trained encoder, and fixed random encoders can be effective for state entropy estimation without having been trained on any data.
We remark that representations from the pre-trained ImageNet encoder could not be useful for our setup, due to the different visual characteristics of natural images in the ImageNet dataset and image observations in our experiments (see Figure~\ref{fig:dm_control_example} for examples of image observations).

\newpage

{\bf Comparison with other exploration methods. }
We also compare our state entropy maximization scheme to other exploration methods combined with RAD, i.e., RAD + RND and RAD + ICM, that learn additional models to obtain intrinsic rewards proportional to prediction errors.
As shown in Figure~\ref{fig:dm_control_exploration}, RAD + RE3 consistently exhibits superior sample efficiency in most tasks.
While RND similarly employs a fixed random network for the intrinsic reward, it also introduces an additional network which requires training and therefore suffers from instability.\footnote{\citet{taiga2019benchmarking} also observed that additional techniques were critical to the performance of RND.}
This result demonstrates that RE3 can improve sample-efficiency without introducing additional models for exploration, by utilizing fixed representations from a random encoder for stable state entropy estimation.

{\bf Compute-efficiency. }
We show that RE3 is a practical and scalable approach for exploration in RL due to its compute-efficiency.
In particular, RE3 is compute-efficient in that (a) there are no gradient updates through the random encoder, and (b) there are no unnecessary forward passes for obtaining representations at every update step since we store low-dimensional latent representations in the replay buffer. 
To evaluate compute-efficiency, we show the floating point operations (FLOPs) consumed by RAD, RAD + SE w/ Random (ours), RAD + SE w/ Contrastive, and RAD + SE w/ Inverse dynamics.
We account only for forward and backward passes through neural network layers.
We explain our full procedure for counting FLOPs in Appendix~\ref{appendix:flop_counting}. 
Figure \ref{fig:dm_control_analysis_compute_efficiency} shows the FLOPS used by each agent to achieve its final performance in Hopper Hop.
One can see that estimating state entropy with a random encoder is significantly more compute-efficient than with a encoder learned by contrastive learning and inverse dynamics prediction.
In particular, RAD + SE w/ Random requires 7.233e+16 FLOPs to achieve its performance at 500K steps, while RAD + SE w/ Inverse dynamics requires roughly twice as many.
One important detail here is that RAD + SE w/ Random has comparable compute-efficiency to RAD.
Therefore, we improve the sample-efficiency of RAD without sacrificing compute-efficiency.

\begin{figure} [t] \centering
\includegraphics[width=0.48\textwidth]{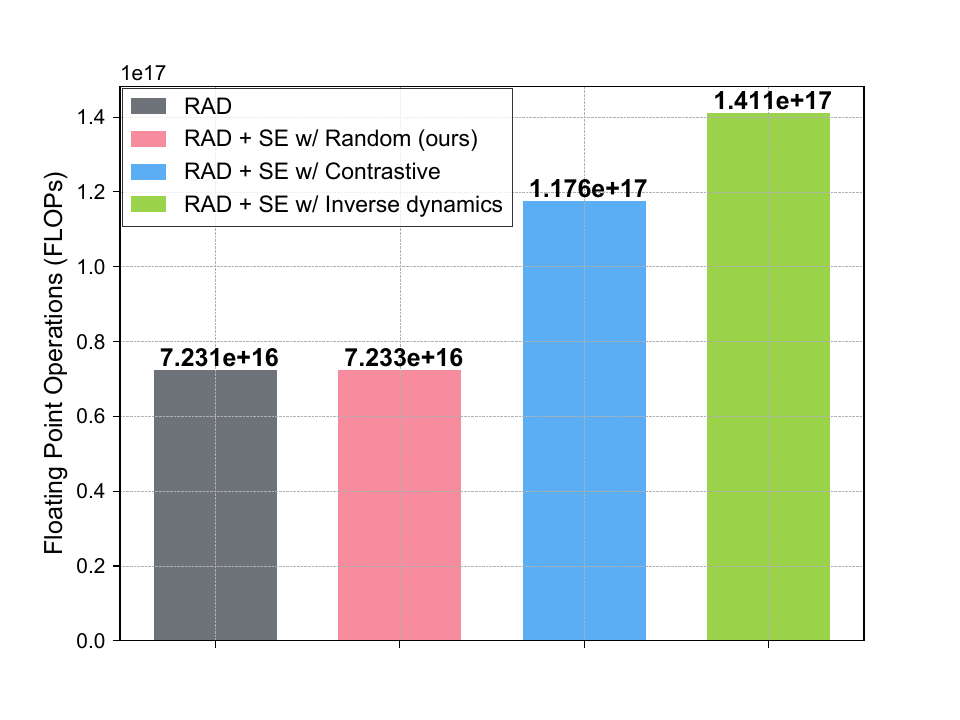}
\vspace{-0.5in}
\caption{Number of FLOPs used by each agent to achieve its performance at 500K environment steps in Hopper Hop (see Figure \ref{fig:dm_control_representation} for corresponding learning curves).}
\label{fig:dm_control_analysis_compute_efficiency}
\vspace{-0.15in}
\end{figure}

\begin{figure*} [ht] \centering
\vspace{-0.125in}
\includegraphics[width=0.97\textwidth]{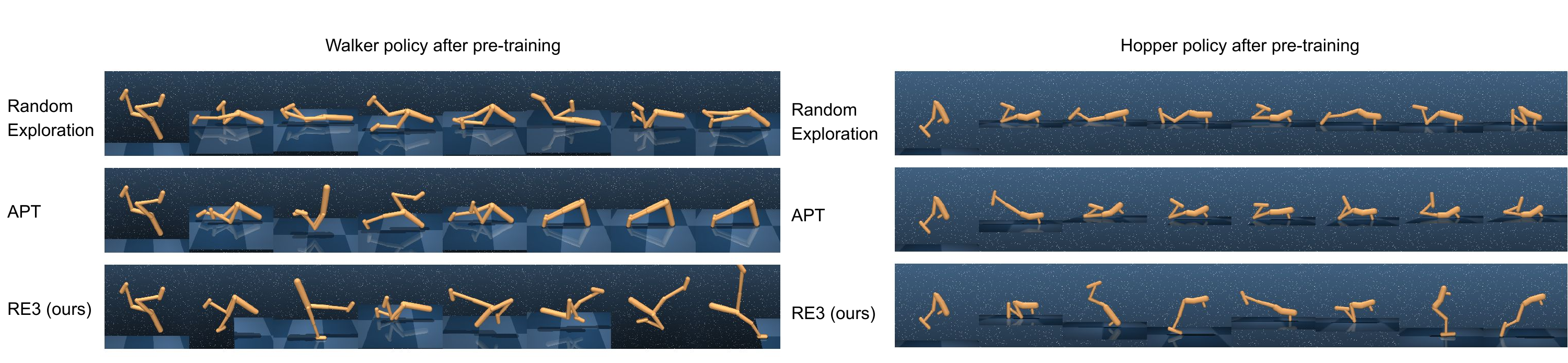}
\vspace{-0.15in}
\caption{Observations from evenly-spaced intervals for one episode of executing policy actions. As a baseline, we show random exploration, i.e. sampling from the action space uniformly at random. We compare the diversity of visited states resulting from pre-training for 500K steps with APT \citep{liu2021behavior} and RE3 (ours). We provide corresponding videos in our website.}
\label{fig:dm_control_behavior_visualization}
\vspace{-0.2in}
\end{figure*}

\begin{figure*} [ht] \centering
\subfigure[Ground-truth state entropy estimate]
{
\includegraphics[width=0.31\textwidth]{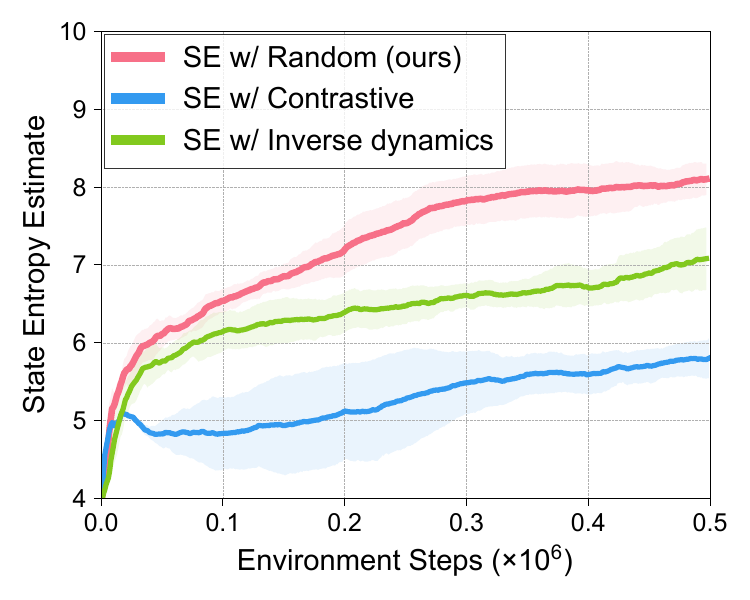}
\label{fig:dm_control_unsupervised_state_entropy}}
\subfigure[Hopper Hop]
{
\includegraphics[width=0.31\textwidth]{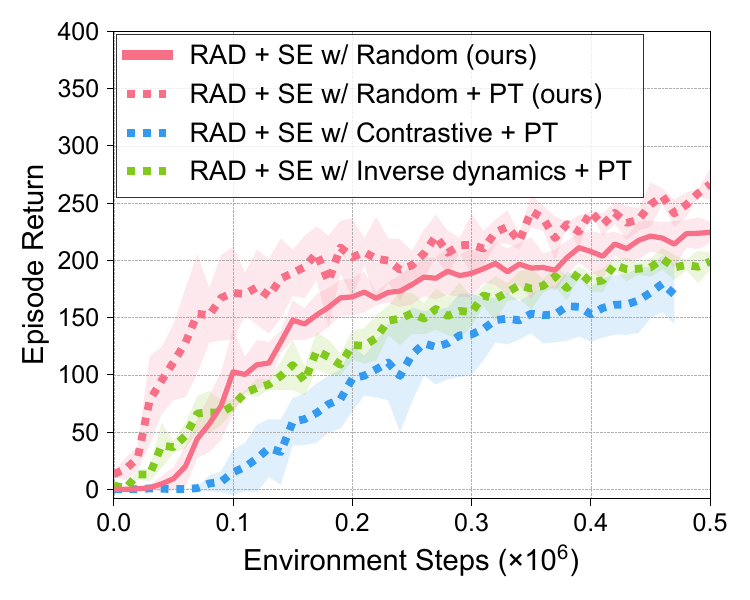}
\label{fig:dm_control_unsupervised_hopper_hop}} 
\subfigure[Hopper Stand]
{
\includegraphics[width=0.31\textwidth]{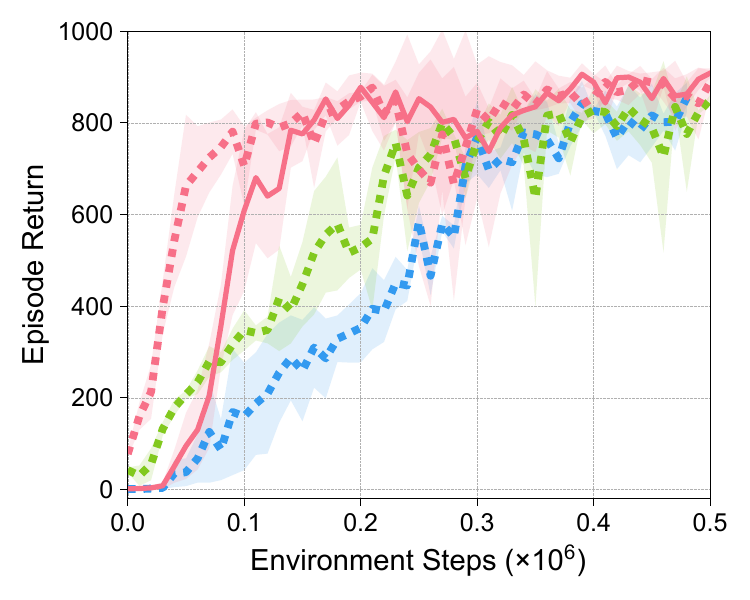}
\label{fig:dm_control_unsupervised_hopper_stand}} 
\vspace{-0.125in}
\caption{
(a) We observe that pre-training Hopper agent with RE3 results in a higher state entropy estimate in the ground-truth state space, i.e., proprioceptive state space, compared to other state entropy maximization schemes that involve representation learning.
This shows that maximizing state entropy in the fixed representation space of a randomly initialized encoder effectively encourages the agent to visit a wide range of states.
We show that this leads to better sample-efficiency when fine-tuning pre-trained policies on (b) Hopper Hop and (c) Hopper Stand.
The solid (or dotted) line and shaded regions represent the mean and standard deviation, respectively, across three runs.}
\label{fig:dm_control_unsupervised}
\vspace{-0.125in}
\end{figure*}

{\bf Evaluation of unsupervised pre-training. }
To evaluate the effectiveness of RE3 for learning diverse behaviors in the pre-training phase without extrinsic rewards, we visualize the behaviors of policies pre-trained for 500K environment steps in Figure~\ref{fig:dm_control_behavior_visualization}.
One can see that the pre-trained policy using RE3 exhibits
more diverse behaviors compared to random exploration or APT \cite{liu2021behavior},
where a policy is pre-trained to maximize state entropy estimate in contrastive representation space.
To further evaluate the diversity of behaviors quantitatively, we show the state entropy estimate in the ground-truth state space of Hopper environment in Figure~\ref{fig:dm_control_unsupervised_state_entropy}, which is computed using distances between all proprioceptive states in the current minibatch.
We observe that RE3 exhibits a higher ground-truth state entropy estimate than state entropy maximization schemes that use contrastive learning and inverse dynamics prediction during pre-training.
This implies that RE3 can effectively maximize the ground-truth state entropy without being able to directly observe underlying ground-truth states.

{\bf Fine-tuning in downstream tasks. }
We also remark that the diversity of behaviors leads to superior sample-efficiency when fine-tuning a pre-trained policy in downstream tasks, as shown in Figure~\ref{fig:dm_control_unsupervised_hopper_hop} and~\ref{fig:dm_control_unsupervised_hopper_stand}.
Specifically, we fine-tune a pre-trained policy in downstream tasks where extrinsic rewards are available, by initializing the parameters of policies with parameters of pre-trained policies (see Appendix~\ref{appendix:dm_control_detail} for more details). We found that fine-tuning a policy pre-trained with RE3 (RAD + SE w/ Random + PT) further improves the sample-efficiency of RAD + SE w/ Random, and also outperforms other pre-training schemes.
We emphasize that RE3 allows learning such diverse behaviors by pre-training a policy only for 500K environment steps, while previous work \cite{liu2021behavior} reported results by training for 5M environment steps.

\begin{figure*}[t!]
    \vspace{-0.1in}
    \centering
    \sbox{\measurebox}{
    \begin{minipage}[b]{.4\textwidth}
    \centering
    \subfigure
      [Robustness to noise by ensemble]
      {\label{fig:robust_noise}\includegraphics[width=\textwidth,height=0.225\textwidth]{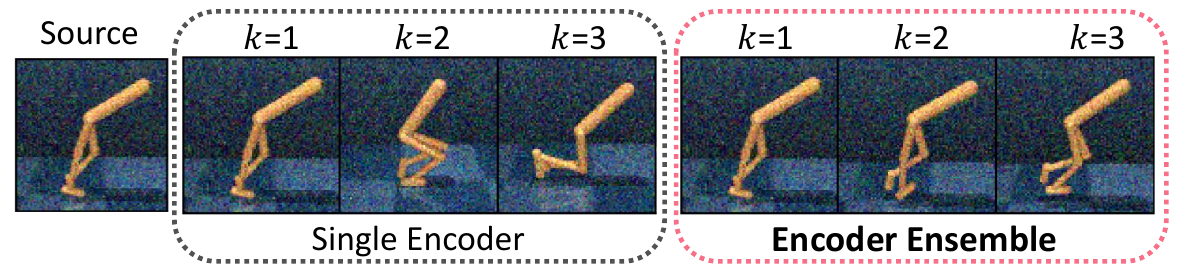}}
    \vfill
    \subfigure
      [Robustness to color changes by ensemble]
      {\label{fig:robust_color}\includegraphics[width=\textwidth,height=0.225\textwidth]{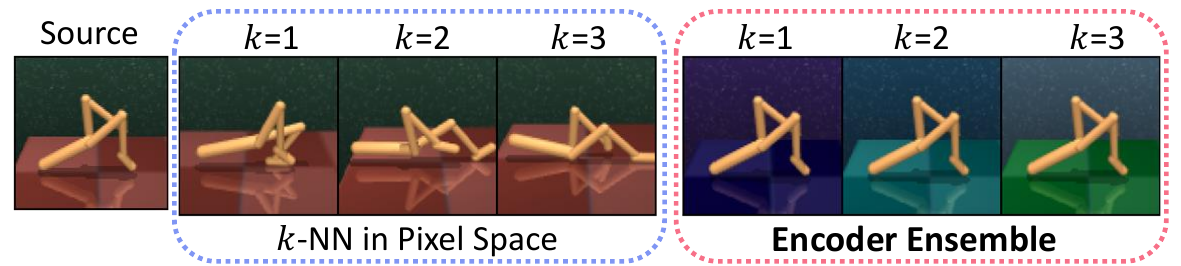}}
    \end{minipage}} 
    \usebox{\measurebox} 
        \begin{minipage}[b][\ht\measurebox][s]{.29\textwidth}
        \subfigure
        [Effects of $k$]
        {\label{fig:ablation_k}\includegraphics[width=\textwidth,height=.8\textwidth]{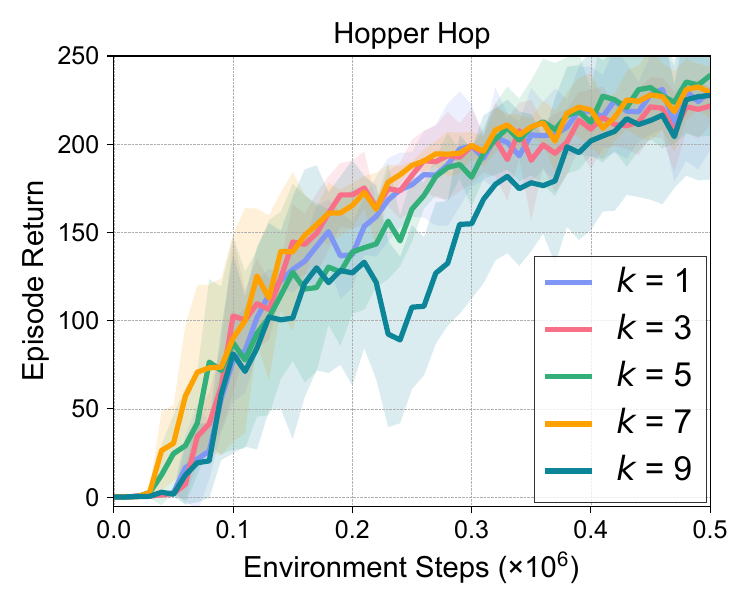}}
        \end{minipage} \hfill
        \begin{minipage}[b][\ht\measurebox][s]{.29\textwidth}
        \subfigure
        [Effects of Initialization]
        {\label{fig:ablation_init}\includegraphics[width=\textwidth,height=.8\textwidth]{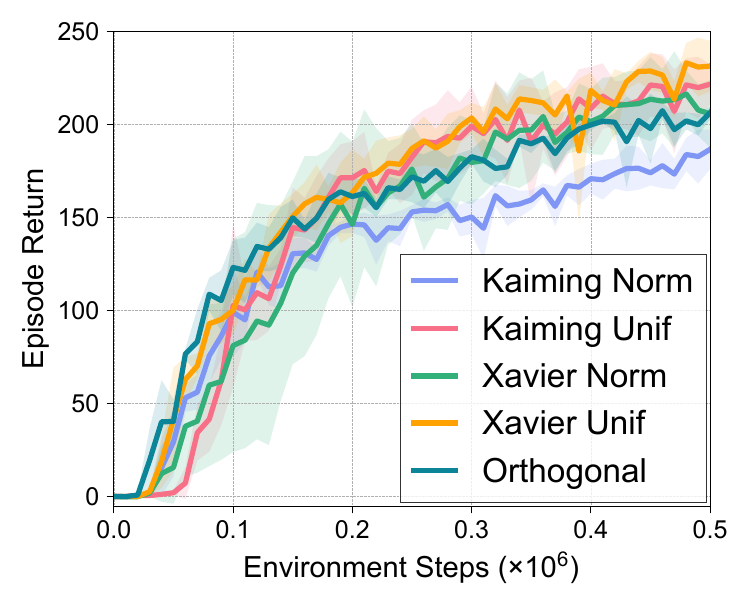}}
        \end{minipage}
\vspace{-0.15in}
\caption{We observe that introducing an ensemble of random encoders can improve robustness to (a) simple Gaussian noise and (b) color changes. Performance of RAD + RE3 with varying (c) $k$ and (d) the initialization of a random encoder on Hopper Hop environment.}
\label{fig:analysis}
\vspace{-0.15in}
\end{figure*}

\begin{figure*} [t!] \centering
\includegraphics[width=0.315\textwidth]{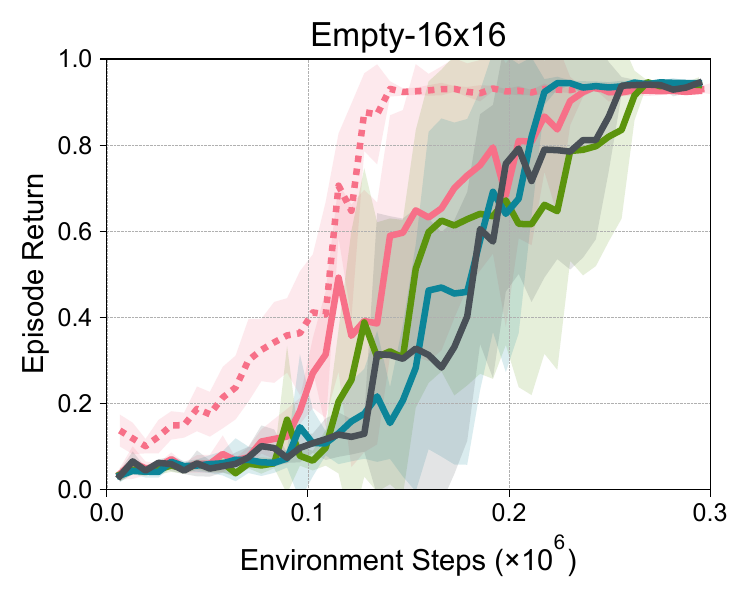}
\includegraphics[width=0.315\textwidth]{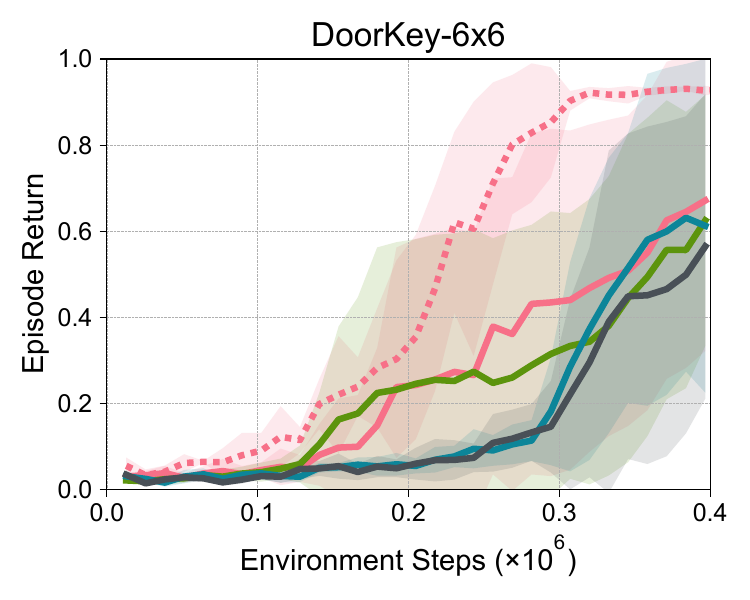}
\includegraphics[width=0.315\textwidth]{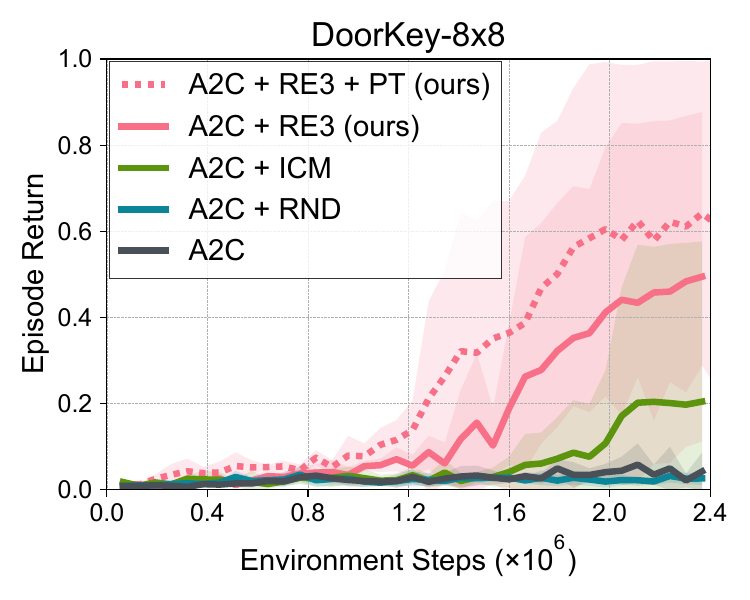}
\vspace{-0.15in}
\caption{Performance on navigation tasks from MiniGrid. A2C + RE3 outperforms other exploration methods in terms of sample-efficiency, and A2C + RE3 + PT further improves sample-efficiency. The solid (or dotted) line and shaded regions represent the mean and standard deviation, respectively, across five runs.}
\label{fig:minigrid_all}
\vspace{-0.15in}
\end{figure*}

{\bf Robustness to noise and perturbations via ensembles. }
We consider a simple extension of our approach by introducing an ensemble of random encoders and found that this improves robustness to simple Gaussian noise see Figure~\ref{fig:robust_noise}. 
We also remark that random encoders can be useful not only for compute-efficiency, but also for $k$-NN selection in more diverse scenarios.
For example, when the background color changes randomly,
one might want to ignore the background color and select $k$-NNs with similar joint positions.
In Figure~\ref{fig:robust_color}, we demonstrate that $k$-NN in raw pixel space only finds observations with similar background colors, while the ensemble of random encoders can find observations with similar joint positions but different colors, as averaging features of convolutional encoders with different initializations could improve robustness to low-level features like colors and textures \citep{lee2019network}.

{\bf Sensitivity analysis to hyperparameters. } 
We investigate how hyperparameters affect the performance of RE3. Specifically, we consider $k \in \{1, 3, 5, 7, 9\}$ for $k$-NN in (\ref{eq:simplied_knn_state_entropy_estimator}), and various initialization schemes for a random encoder, i.e., the Xavier initialization (also called the Glorot initialization; \citealt{glorot2010understanding}), the He initialization \cite{he2015delving}, and the Orthogonal initialization \cite{saxe2013exact}. Figure~\ref{fig:ablation_k} and Figure~\ref{fig:ablation_init} show that RE3 is robust to such considered hyperparameters.

\subsection{MiniGrid Experiments}
\begin{figure} \centering
\vspace{-0.25in}
\includegraphics[width=0.45\textwidth]{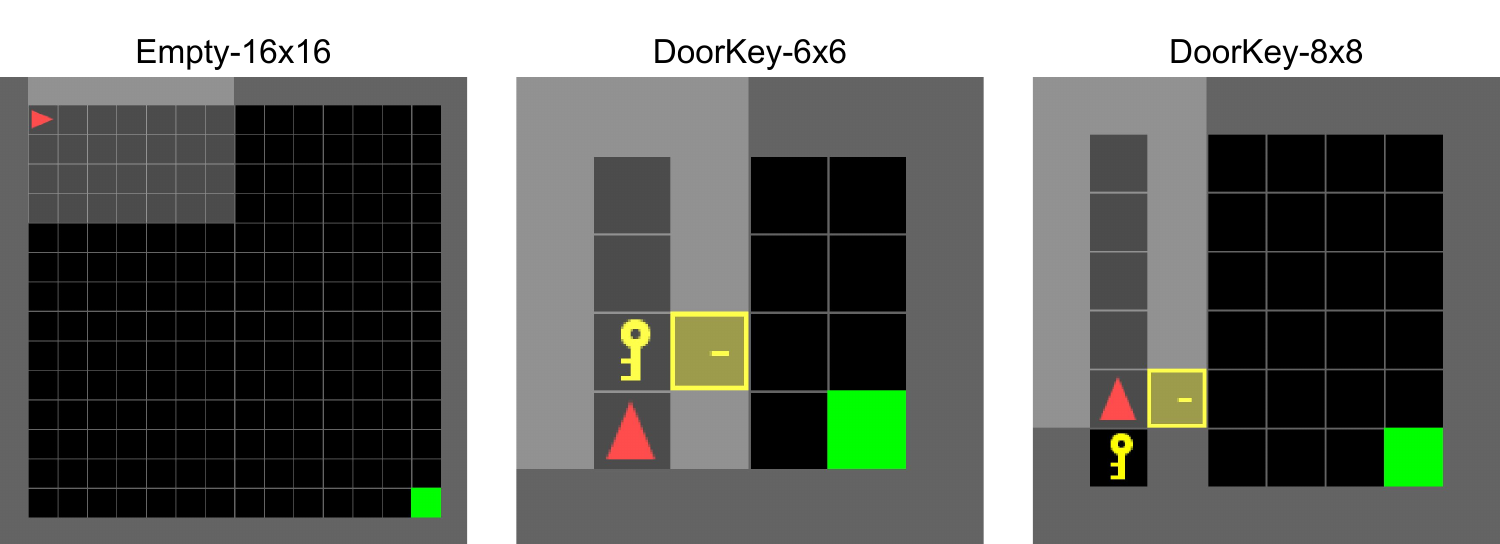}
\vspace{-0.08in}
\caption{Navigation tasks from MiniGrid \citep{gym_minigrid} used in our experiments. The agent is represented as a red arrow and the light gray region shows the $7 \times 7$ (or smaller, if obstructed by walls) grid which the agent observes. The agent receives a positive reward only for reaching the green square.}
\label{fig:minigrid_grids}
\vspace{-0.2in}
\end{figure}

{\bf Setup. }
We evaluate our method on MiniGrid \citep{gym_minigrid}, a gridworld environment with a selection of sparse reward tasks. We consider the following setups where the agent obtains a reward only by reaching the green goal square: Empty, a large room with the goal in the furthest corner; DoorKey, where the agent must collect a key and unlock a door before entering the room containing the goal. The tasks are shown in Figure \ref{fig:minigrid_grids}. For evaluation, we consider two exploration methods, RND and ICM.
For our method and other exploration methods, we use Advantage Actor-Critic (A2C; \citet{mnih2016asynchronous}) as the underlying RL algorithm.
In all tasks, the agent has access to a compact $7 \times 7 \times 3$ embedding of the $7 \times 7$ grid directly in front of it, making the environment partially-observable. To combine RE3 with A2C, an on-policy RL method, we maintain a replay buffer of 10K samples solely for computing the RE3 intrinsic reward, and compute $k$-NN distances between the on-policy batch and the entire replay buffer. For RND and ICM, the intrinsic reward is computed using the on-policy batch. For RE3, ICM, and RND, we perform hyperparameter search over the intrinsic reward weight and report the best result (see the Appendix~\ref{appendix:minigrid_detail} for more details).

\begin{figure*} [t!] \centering
\vspace{-0.15in}
\subfigure[Montezuma's Revenge]
{
\includegraphics[width=0.31\textwidth]{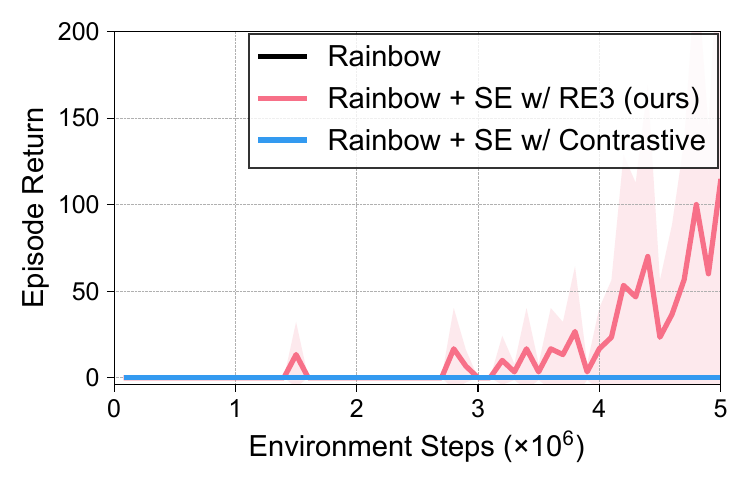}
\label{fig:atari_main_montezuma}}
\subfigure[Beam Rider]
{
\includegraphics[width=0.31\textwidth]{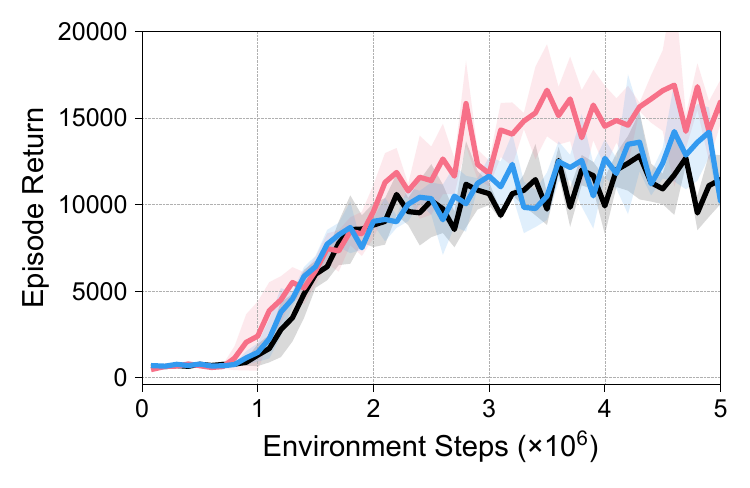}
\label{fig:atari_main_beam_rider}} 
\subfigure[Human Normalized Score]
{
\includegraphics[width=0.31\textwidth]{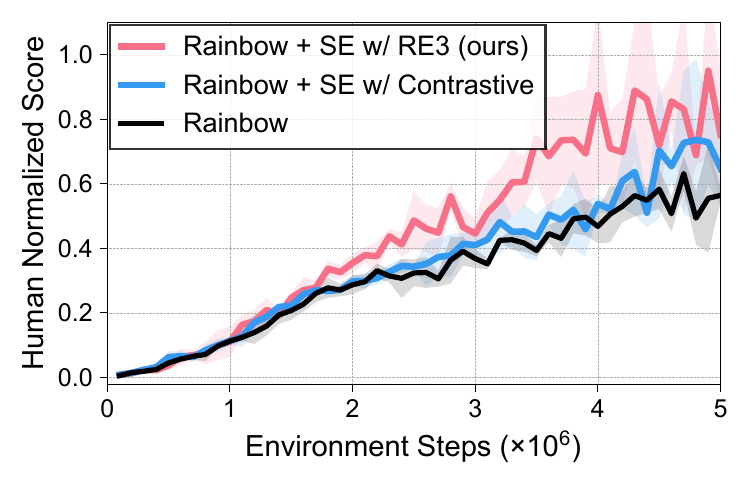}
\label{fig:atari_hns}}
\vspace{-0.15in}
\caption{
Performance on (a) Montezuma's Revenge and (b) Beam Rider games. (c) Human normalized score averaged over six Atari games. The solid line and shaded regions represent the mean and standard deviation, respectively, across three runs.
}
\label{fig:atari_results}
\vspace{-0.15in}
\end{figure*}

{\bf Comparison with other exploration methods. }
Figure~\ref{fig:minigrid_all} shows that RE3 is more effective for improving the sample-efficiency of A2C in most tasks, compared to other exploration methods, RND and ICM, that learn additional models.
In particular, A2C + RE3 achieves average episode return of 0.49 at 2.4M environment steps in DoorKey-8x8; in comparison, A2C + ICM achieves a return of 0.20 and A2C + RND and A2C both fail to achieve non-trivial returns.
These results demonstrate that state entropy maximization with RE3 can also improve the sample-efficiency of on-policy RL algorithms by introducing only a small-size replay buffer.

{\bf Fine-tuning in downstream tasks. } To evaluate the effectiveness of RE3 for unsupervised pre-training in MiniGrid tasks, we first pre-train a policy in a large vacant room (Empty-16$\times$16) to maximize RE3 intrinsic rewards for 100K environment steps.
Then, we fine-tune the pre-trained policy in downstream tasks by initializing a policy with pre-trained parameters and subsequently training with A2C + RE3.
Figure~\ref{fig:minigrid_all} shows that A2C + RE3 + PT significantly improves the sample-efficiency of A2C, which demonstrates that the ability to explore novel states in a large empty room helps improve sample-efficiency in DoorKey tasks, which involve the added complexity of additional components, e.g., walls, doors, and locks. We show a comparison to state entropy maximization with contrastive learning in Figure~\ref{fig:minigrid_with_apt} and observe that it does not work well, as contrastive learning depends on data augmentation specific to images (e.g., random shift and color jitter), which are not compatible with the compact embeddings used as inputs for MiniGrid.
RE3 effectively eliminates the need for carefully chosen data augmentations by employing a random encoder.

\subsection{Atari Experiments}
We also evaluate RE3 on Atari games from Arcade Learning Environment \cite{bellemare2013arcade}.
We use Rainbow \cite{hessel2018rainbow} as the underlying RL algorithm, and use convolutional neural networks with the same architecture as in Rainbow for random encoders.
For evaluation, we perform hyperparameter search over the intrinsic reward weight for each environment, and report the human normalized score \cite{mnih2015human} over six Atari games (see Appendix~\ref{appendix:atari_detail} for more details).
Figure~\ref{fig:atari_results} shows that RE3 exhibits superior sample-efficiency compared to Rainbow and Rainbow + SE w/ Contrastive on various Atari games, including hard exploration games like Montezuma's revenge (see Figure~\ref{fig:atari_supple} for additional experimental results).
These results demonstrate that random encoders can also be useful in more visually complex environments.

\section{Discussion}
In this paper, we present RE3, a simple exploration method compatible with both model-free and model-based RL algorithms.
RE3 maximizes a $k$-nearest neighbor state entropy estimate in the fixed representation space of a randomly initialized encoder, which effectively captures information about similarity between states without any training. Our experimental results demonstrate that RE3 can encourage exploration in
widely-used
benchmarks, as it enables stable and compute-efficient state entropy estimation.
Here, we emphasize that our goal is not to claim that representation learning or additional models are not required for exploration, but to show that fixed random encoders can be useful for efficient exploration.
For more visually complex domains, utilizing pre-trained fixed representations for stable state entropy estimation could be more useful,
but we leave it to future work to explore this direction further because this would require having access to environments and a wide distribution of states for pre-training, which is itself a non-trivial problem.
Another interesting direction would be to investigate the effect of network architectures for state entropy estimation, or to utilize state entropy for explicitly guiding the action of a policy to visit diverse states.
We believe RE3 would facilitate future research by providing a simple-to-implement, stable, and compute-efficient module that can be easily combined with other techniques.

\section*{Acknowledgements}
This research is supported in part by Open Philanthropy,
ONR PECASE N000141612723, 
NSF NRI \#2024675,
Tencent, Berkeley Deep Drive,
NSF IIS \#1453651,
Engineering Research Center Program through the National Research Foundation of Korea (NRF) funded by the Korean Government MSIT (NRF-2018R1A5A1059921),
and Institute of Information \& communications Technology Planning \& Evaluation (IITP) grant funded by the Korea government (MSIT) (No.2019-0-00075, Artificial Intelligence Graduate School Program (KAIST)).
We would like to thank Seunghyun Lee, Kibok Lee, Colin Li, Sangwoo Mo, Mandi Zhao, and Aravind Srinivas for providing helpful feedbacks and suggestions.
We also appreciate Hao Liu for providing implementation details, and Adam Stooke for providing pre-trained parameters of ATC.

\bibliography{main}
\bibliographystyle{icml2021}

\appendix
\onecolumn

\begin{center}{\bf {\LARGE Appendix}}
\end{center}

\section{Details on DeepMind Control Suite Experiments}
\label{appendix:dm_control_detail}

\subsection{Environments}
We evaluate the performance of RE3 on various tasks from DeepMind Control Suite \cite{tassa2020dm_control}. For Hopper Hop, Quadruped Run, Cartpole Swingup Sparse, Pendulum Swingup, we use the publicly available environments without any modification. For environments which are not from the publicly available released implementation repository (\url{https://github.com/deepmind/dm_control}), we designed the tasks following \citet{seyde2021learning} as below:
\begin{itemize}[topsep=0.5pt,leftmargin=5.5mm]
    \item \textbf{Walker/Cheetah Run Sparse}: The goal of Walker/Cheetah Run Sparse task is same as in Walker/Cheetah Run, moving forward as fast as possible, but reward is given sparsely until it reaches a certain threshold: $r = r_{\tt{original}} \cdot \mathds{1}_{r_{\tt{original}} > 0.25}$, where $r_{\tt{original}}$ is the reward in Walker/Cheetah Run from DeepMind Control Suite.
\end{itemize}

\subsection{Implementation Details for Model-free RL} \label{app:setups}
For all experimental results in this work, we report the results obtained by running experiments using the publicly available released implementations from the authors\footnote{We found that there are some differences from the reported results in the original papers which are due to different random seeds.
We provide full source code and scripts for reproducing the main results to foster reproducibility.}, i.e., RAD (\url{https://github.com/MishaLaskin/rad}) and DrQ (\url{https://github.com/denisyarats/drq}).
We use random crop augmentation for RAD and random shift augmentation for DrQ.
We provide a full list of hyperparameters in Table~\ref{tbl:rad_re3_hyperparameters}.

{\bf Implementation details for RE3. } We highlight key implementation details for RE3:
\begin{itemize}[topsep=0.5pt,leftmargin=5.5mm]
    \item \textbf{Intrinsic reward. } We use $r^{\tt{i}}(s_{i}) := ||y_{i} - y^{\text{$k$-NN}}_{i}||_{2}$ for the intrinsic reward. We got rid of $\log$ from intrinsic reward in (\ref{eqn:re3_intrinsic}) for simplicity in DeepMind Control Suite experiments, but results using $\log$ are also similar. To make the scale of intrinsic reward $r^{\tt{i}}$ consistent across tasks, following \citet{liu2021behavior}, we normalize the intrinsic reward by dividing it by a running estimate of the standard deviation. As for newly introduced hyperparameters, we use $k=3$, and perform hyperparameter search over $\beta_{0} \in \{0.05, 0.25\}$ and $\rho \in \{0.00001, 0.000025\}$.
    \item \textbf{Architecture. } For all model-free RL methods, we use the same encoder architecture as in \citet{yarats2019improving}. Specifically, this encoder consists of 4 convolutional layers followed by ReLU activations. We employ kernels of size 3 $\times$ 3 with 32 channels for all layers, and 1 stride except of the first layer which has stride 2. The output of convolutional layers is fed into a single fully-connected layer normalized by LayerNorm \cite{ba2016layer}. Finally, tanh nonlinearity is added to the 50-dimensional output of the fully-connected layer.
    \item \textbf{Unsupervised pre-training. } For unsupervised pre-training, we first train a policy to maximize intrinsic rewards in (\ref{eqn:re3_intrinsic}) without extrinsic rewards for 500K environment steps.
    For fine-tuning a policy in downstream tasks, we initialize the parameters using the pre-trained parameters and then learn a policy to maximize RE3 objective in (\ref{eqn:combined_reward}) for 500K environment steps. To stabilize the initial fine-tuning phase by making the scale of intrinsic reward consistent across pre-training and fine-tuning, we load the running estimate of the standard deviation from pre-training phase.
\end{itemize}

{\bf Implementation details for representation learning baselines . } We highlight key implementation details for state entropy maximization schemes that involve representation learning, i.e., contrastive learning and inverse dynamics prediction, and that employs a pre-trained ImageNet Encoder:
\begin{itemize}[topsep=0.5pt,leftmargin=5.5mm]
    \item \textbf{Contrastive learning. }
    For state entropy maximization with contrastive learning, we introduce a separate convolutional encoder with randomly initialized parameters learned to minimize contrastive loss \cite{srinivas2020curl}. Note that we use the separate encoder to investigate the independent effect of representation learning.
    Then we compute intrinsic reward $r^{\tt{i}}$ using representations obtained by processing observations through this separate encoder.
    \item \textbf{Inverse dynamics prediction. } For state entropy maximization with inverse dynamics prediction, we introduce a separate convolutional encoder with randomly initialized parameters learned to predict actions from two consecutive observations. Specifically, we introduce 2 fully-connected layers with ReLU activations to predict actions on top of encoder representations. We remark that this separate encoder is separate from RL. Then we compute intrinsic reward $r^{\tt{i}}$ using representations obtained by processing observations through this separate encoder.
    \item \textbf{Pre-trained ImageNet encoder. } For state entropy maximization with pre-trained ImageNet encoder, we utilize a ResNet-50 \cite{he2016deep} encoder from publicly available torchvision models (\url{https://pytorch.org/vision/0.8/models.html}). To compute intrinsic reward $r^{\tt{i}}$, we utilize representations obtained by processing observations through this pre-trained encoder.
    \item \textbf{Pre-trained ATC encoder. } For state entropy maximization with pre-trained ATC \citep{stooke2020decoupling} encoder, we utilize pre-trained encoders from the authors. Specifically, these encoders are pre-trained by contrastive learning on pre-training datasets that contain samples encountered while training a RAD agent on DeepMind Control Suite environments (see \citet{stooke2020decoupling} for more details). We use pre-trained parameters from Walker Run, Hopper Stand, and Cheetah Run for RAD + SE w/ ATC on Walker Run Sparse, Hopper Hop, and Cheetah Run Sparse, respectively. To compute intrinsic reward $r^{\tt{i}}$, we utilize representations obtained by processing observations through this pre-trained encoder.
\end{itemize}

{\bf Implementation details for exploration baselines. } We highlight key implementation details for exploration baselines, i.e., RND \cite{burda2018exploration} and ICM \cite{pathak2017curiosity}:
\begin{itemize}[topsep=0.5pt,leftmargin=5.5mm]
    \item \textbf{RND. } For RND, we introduce a random encoder $f_{\theta}$ whose architecture is same as in \citet{yarats2019improving}, and introduce a predictor network $g_{\phi}$ consisting of a convolutional encoder with the same architecture and 2-layer fully connected network with 1024 units each. Then, parameters $\phi$ of the predictor network are trained to predict representations from a random encoder given the same observations, i.e., minimize $\epsilon = ||f_{\theta}(s_{i}) - g_{\phi}(s_{i})||_{2}$. We use prediction error $\epsilon$ as an intrinsic reward and learn a policy that maximizes $r^{\tt{total}} = r^{\tt{e}} + \beta \cdot r^{\tt{i}}$. We perform hyperparameter search over the weight $\beta \in \{0.05, 0.1, 1.0, 10.0\}$ and report the best result on each environment.
    \item \textbf{ICM. } For ICM, we introduce a convolutional encoder $g_{\phi}$ whose architecture is same as in \citet{yarats2019improving}, and introduce a inverse dynamics predictor $h_{\psi}$ with 2 fully-connected layers with 1024 units. These networks are learned to predict actions between two consecutive observations, i.e., minimize $\mathcal{L}_{\tt{inv}} = ||a_{t} - h_{\psi}(g_{\phi}(s_{t}), g_{\phi}(s_{t+1}))||_{2}$. We also introduce a forward dynamics predictor $f_{\Psi}$ that is learned to predict the representation of next time step, i.e., minimize $\mathcal{L}_{\tt{forward}} = \frac{1}{2}||g_{\phi}(s_{t+1}) - f_{\Psi}(g_{\phi}(s_{t}), a_{t})||^{2}_{2}$. Then, we use prediction error as an intrinsic reward and learn a policy that maximizes $r^{\tt{total}} = r^{\tt{e}} + \beta \cdot r^{\tt{i}}$. For joint training of the forward and inverse dynamics, following \citet{pathak2017curiosity}, we minimize $0.2 \cdot \mathcal{L}_{\tt{forward}} + 0.8 \cdot \mathcal{L}_{\tt{inv}}$. We perform hyperparameter search over the weight $\beta \in \{0.05, 0.1, 1.0\}$ and report the best result on each environment. 
\end{itemize}

\subsection{Implementation Details for Model-based RL. }
For Dreamer, we use the publicly available released implementation repository (\url{https://github.com/danijar/dreamer}) from the authors\footnote{We use the newer implementation of Dreamer following the suggestion of the authors, but we found that there are some difference from the reported results in the original paper which are might be due to the difference between newer implementation and the original implementation, or different random seeds. We provide full source code and scripts for reproducing the main results to foster reproducibility.}.
We highlight key implementation details for the combination of Dreamer with RE3:
\begin{itemize}[topsep=0.5pt,itemsep=0.85pt]
    \item \textbf{Intrinsic reward. } We use $r^{\tt{i}}(s_{i}) := ||y_{i} - y^{\text{$k$-NN}}_{i}||_{2}$ for the intrinsic reward. We got rid of $\log$ from intrinsic reward in (\ref{eqn:re3_intrinsic}) for simplicity in DeepMind Control Suite experiments, but results using $\log$ are also similar. To make the scale of intrinsic reward $r^{\tt{i}}$ consistent across tasks, following \citet{liu2021behavior}, we normalize the intrinsic reward by dividing it by a running estimate of the standard deviation. Since Dreamer utilizes trajectory segments for training batch, we use large value of $k = 50$ to avoid find $k$-NN only within a trajectory segment. We also use $\beta_{0} = 0.1$, and $\rho = 0.0$, i.e., no decay schedule. We also remark that we find $k$-NN only within a minibatch instead of the entire buffer as in model-free RL, since the large batch size of Dreamer is enough for stable entropy estimation.
    \item \textbf{Architecture. } We use the same convolutional architecture as in Dreamer. Specifically, this encoder consists of 4 convolutional layers followed by ReLU activations. We employ kernels of size 4 $\times$ 4 with $\{32, 64, 128, 256\}$ channels, and 2 stride for all layers. To obtain low-dimensional representations, we additionally introduce a fully-connected layer normalized by LayerNorm~\cite{ba2016layer}. Finally, tanh nonlinearity is added to the 50-dimensional output of the fully-connected layer.
\end{itemize}

\clearpage

\begin{table}[h]
\caption{Hyperparameters of RAD + RE3 used for DeepMind Control Suite experiments.}
\vskip 0.15in
\begin{center}
\begin{tabular}{ll}
\toprule
\textbf{Hyperparameter} & \textbf{Value}  \\
\midrule
Augmentation    & Crop  \\ 
Observation rendering    & $(100, 100)$  \\
Observation downsampling    & $(84, 84)$  \\ 
Replay buffer size   & $100000$ \\
Initial steps   & $10000$ quadruped, run; $1000$ otherwise \\
Stacked frames    & $3$  \\ 
Action repeat    & $4$ quadruped, run; $2$ otherwise \\
Learning rate (actor, critic) & $0.0002$ \\
Learning rate ($\alpha$) & $0.001$ \\
Batch size & $512$ \\
$Q$ function EMA $\tau$ & $0.01$ \\
Encoder EMA $\tau$ & $0.05$ \\
Critic target update freq & $2$ \\
Convolutional layers & $4$ \\
Number of filters & $32$ \\
Latent dimension & $50$ \\
Initial temperature & $0.1$ RAD + RE3; $0.01$ RAD + RE3 + PT \\
$k$ & 3 \\
Initial intrinsic reward scale $\beta_{0}$ & $0.25$ pendulum, swingup; $0.05$ otherwise \\
Intrinsic reward decay $\rho$ & $0.000025$ walker, run sparse; $0.00001$ otherwise \\
\midrule
Intrinsic reward scale (RND) & $10.0$ cartpole, swingup sparse \\
& $0.1$ pendulum, swingup; cheetah, run sparse \\
& $0.05$ otherwise\\
Intrinsic reward scale (ICM) & $1.0$ cheetah, run sparse\\
& $0.1$ otherwise \\
\bottomrule
\end{tabular}
\label{tbl:rad_re3_hyperparameters}
\end{center}
\vskip -0.1in
\end{table} 

\begin{table}[h]
\caption{Hyperparameters of Dreamer + RE3 used for DeepMind Control Suite experiments. We only specify hyperparameters different from original paper of \citet{hafner2019dream}.}
\vskip 0.15in
\begin{center}
\begin{small}
\begin{tabular}{ll}
\toprule
\textbf{Hyperparameter} & \textbf{Value}  \\
\midrule
Initial episodes & $5$ \\
Precision\footnotemark & $32$ \\
Latent dimension of a random encoder & 50 \\
$k$ & 53 \\
Initial reward scale $\beta_{0}$ & 0.1 \\
Intrinsic reward decay $\rho$ & $0.0$ \\
\bottomrule
\end{tabular}
\label{tbl:dreamer_re3_hyperparameters}
\end{small}
\end{center}
\vskip -0.1in
\end{table} 

\footnotetext{We found that precision of 32 is necessary for avoiding NaN in intrinsic reward normalization, as running estimate of standard deviation is very small.}

\clearpage

\section{Additional Experimental Results on DeepMind Control Suite}
\label{appendix:dm_control_additional_results}
We provide additional experimental results on various tasks from DeepMind Control Suite \cite{tassa2020dm_control}. 
We observe that RE3 improves sample-efficiency in several tasks, e.g., Reacher Hard and Hopper Stand, while not degrading the performance in dense-reward tasks like Cartpole Balance. This demonstrates the simple and wide applicability of RE3 to various tasks.

\begin{figure*} [h!] \centering
\includegraphics[width=0.98\textwidth]{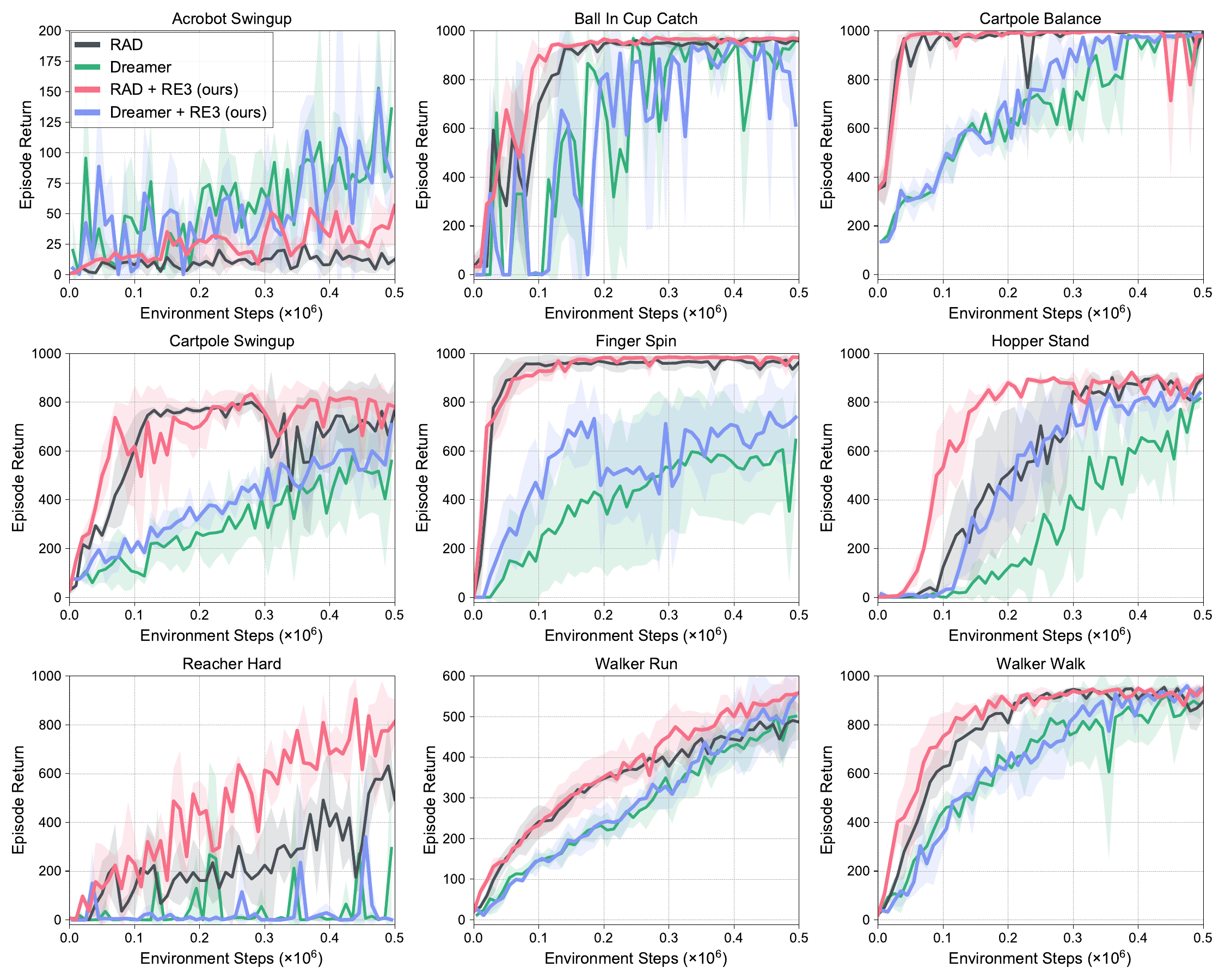}
\caption{Performance on locomotion tasks from DeepMind Control Suite. The solid line and shaded regions represent the mean and standard deviation, respectively, across five runs.}
\label{fig:dm_control_online_supple}
\end{figure*}

\clearpage

\section{Details on MiniGrid Experiments}
\label{appendix:minigrid_detail}
For our A2C implementation in MiniGrid, we use the publicly available released implementation repository (\url{https://github.com/lcswillems/rl-starter-files}) and use their default hyperparameters. We provide a full list of hyperparameters that are introduced by our method or emphasized for clarity in Table \ref{tbl:minigrid_hyperparameters}. 

We highlight some key implementation details:
\begin{itemize}[topsep=0.5pt,leftmargin=5.5mm]
    \item We use $r^{\tt{i}}(s_{i}) := \log(||y_{i} - y^{\text{$k$-NN}}_{i}||_{2} + 1)$ for the intrinsic reward. The additional $1$ is for numerical stability.
    \item We use the average distance between $y_{i}$ and its $k$ nearest neighbors (i.e., $y^{\text{$2$-NN}}_{i}, \cdots , y^{\text{$k$-NN}}_{i}$) for the intrinsic reward, instead of the single $k$ nearest neighbor. This provides a less noisy state entropy estimate and empirically improves performance in MiniGrid environments.
    \item We perform hyperparameter search over $\beta \in \{0.00001, 0.00005, 0.0001, 0.0005, 0.001, 0.005, \\ 0.01, 0.05, 0.1\}$ for RE3, ICM, and RND and report the best result.
    \item We do not change the network architecture from the above publicly available implementation. We use the same encoder architecture as the RL encoder for state entropy maximization. This encoder architecture consists of 3 convolutional layers with kernel 2, stride 1, and padding 0, each followed by a ReLU layer. The convolutional layers have 16, 32, and 64 filters respectively. The first ReLU is followed by a two-dimensional max pooling layer with kernel 2. The actor and critic MLPs both contain two fully-connected layers with hidden dimension 64, with a tanh operation in between. 
\end{itemize}

\begin{table*}[ht]
\caption{Hyperparameters used for MiniGrid experiments. Most hyperparameter values are unchanged across environments with the exception of evaluation frequency and intrinsic reward weight $\beta$.}
\vskip 0.1in
\begin{center}
\begin{small}
\begin{tabular}{ll}
\toprule
\textbf{Hyperparameter} & \textbf{Value}  \\
\midrule
Input Size    & $(7,7,3)$  \\ 
Replay buffer size (for RE3 intrinsic reward)   & 10000 \\ 
Stacked frames    & $1$  \\ 
Action repeat    & $1$ \\
Evaluation episodes    & $100$  \\ 
Optimizer & RMSprop \\
$k$ & 3 \\
Evaluation frequency    & $6400$ Empty-16x16; $12800$ DoorKey-6x6 \\
& $64000$ DoorKey-8x8 \\
Intrinsic reward weight $\beta$ in Empty-16x16 experiments & $0.1$ A2C + RE3 + PT \\
 & $0.1$ A2C + RE3  \\
 & $0.00001$ A2C + ICM \\
 & $0.00005$ A2C + RND \\
Intrinsic reward weight $\beta$ in DoorKey-6x6 experiments & $0.05$ A2C + RE3 + PT \\
 & $0.005$ A2C + RE3  \\
 & $0.0001$ A2C + ICM \\
 & $0.0001$ A2C + RND \\
Intrinsic reward weight $\beta$ in DoorKey-8x8 experiments & $0.05$ A2C + RE3 + PT \\
 & $0.01$ A2C + RE3  \\
 & $0.001$ A2C + ICM \\
 & $0.00005$ A2C + RND \\
Intrinsic reward decay $\rho$ & $0$ \\
Number of processes & $16$ \\
Frames per process  & $5$ \\
Discount $\gamma$ & 0.99 \\
GAE $\lambda$ & 0.95 \\
Entropy coefficient & $0.01$ \\
Value loss term coefficient & $0.5$ \\
Maximum norm of gradient & $0.5$ \\
RMSprop $\epsilon$ & $0.01$ \\
Clipping $\epsilon$ & $0.2$ \\
Recurrence & None \\
\bottomrule
\end{tabular}
\end{small}
\label{tbl:minigrid_hyperparameters}
\end{center}
\vskip -0.1in
\end{table*} 

\newpage

\section{Details on Atari Experiments}
\label{appendix:atari_detail}
For our Rainbow implementation in Atari games, we use the publicly available implementation repository (\url{https://github.com/Kaixhin/Rainbow}) and use their default hyperparameters. 

We highlight some key implementation details:
\begin{itemize}[topsep=0.5pt,leftmargin=5.5mm]
    \item We use $r^{\tt{i}}(s_{i}) := \log(||y_{i} - y^{\text{$k$-NN}}_{i}||_{2} + 1)$ for the intrinsic reward. The additional $1$ is for numerical stability.
    \item We perform hyperparameter search over $\beta \in \{0.0001, 0.001, 0.01\}$ for Rainbow + SE w/ RE3 and report the best result for each environment. We used $0.01$ for Asterix and BeamRider, $0.001$ for Seaquest, and $0.0001$ for other games.
    \item We perform hyperparameter search over $\beta \in \{0.0001, 0.001\}$ for Rainbow + SE w/ Contrastive and report the best result for each environment. Specifically, we used $0.001$ for Seaquest, $0.0001$ for other games.
    \item We do not change the network architecture from the above publicly available implementation. We use the same encoder architecture as the RL encoder for state entropy maximization, with additional linear layer to reduce the dimension of latent representations. This encoder consists of 3 convolutional layers with kernels $\{8, 4, 3\}$, strides $\{4, 4, 3\}$, and padding 0, each followed by a ReLU layer. The convolutional layers have 32, 32, 64 filters, respectively. Then the outputs from a convolutional encoder is flattened and followed by a linear layer of size $\{50, 150\}$. We find that large dimension of 150 is effective for Montezuma's Revenge, but 50 is enough for all other environments.
\end{itemize}

\section{Additional Experimental Results on Atari Games}
\begin{figure*} [h!] \centering
\includegraphics[width=0.98\textwidth]{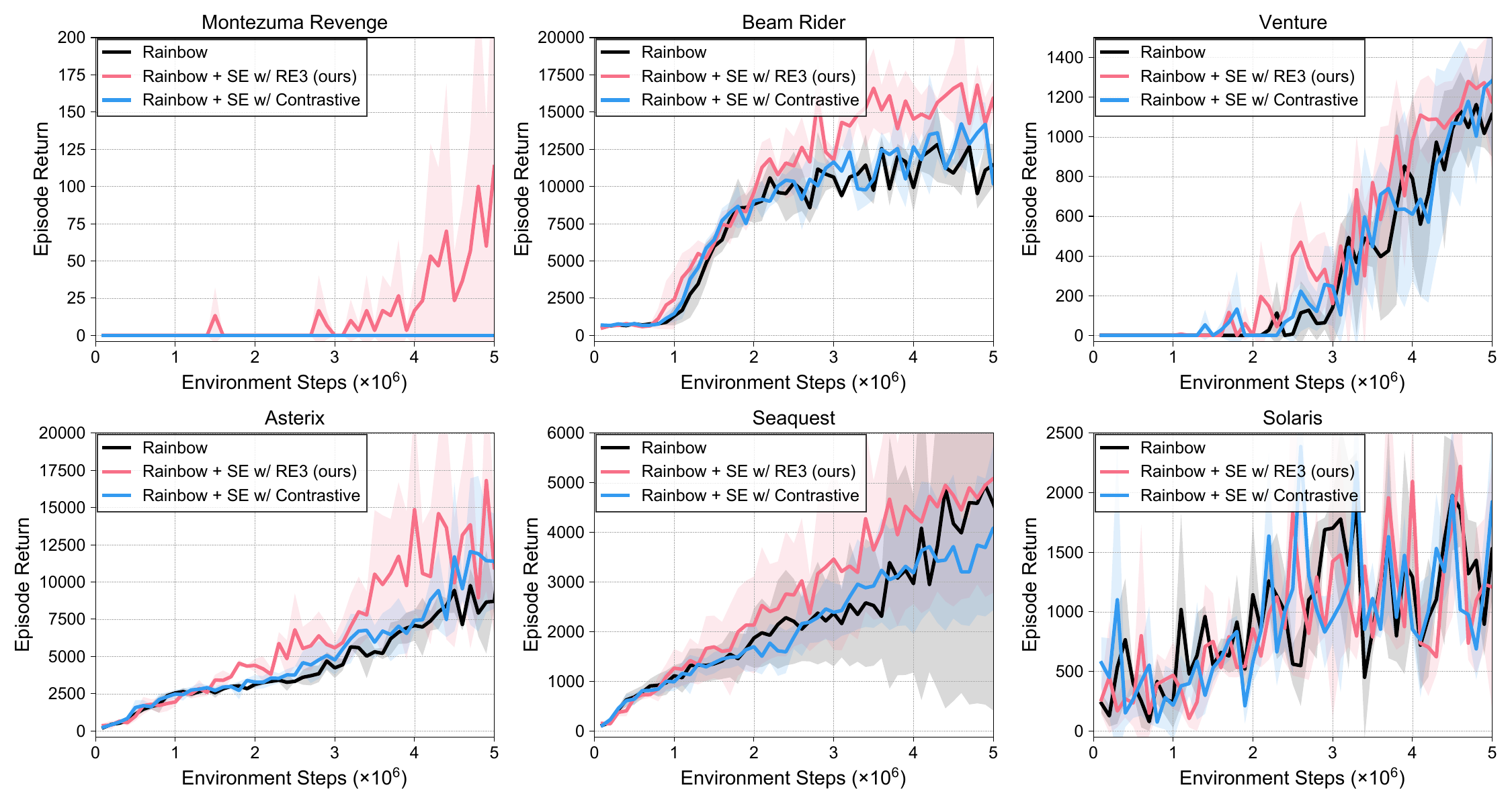}
\caption{Performance on Atari games from Arcade Learning Environment. The solid line and shaded regions represent the mean and standard deviation, respectively, across three runs.}
\label{fig:atari_supple}
\end{figure*}

\newpage

\section{Calculation of Floating Point Operations} \label{appendix:flop_counting}
We explain our FLOP counting procedure for comparing the compute-efficiency of RAD, RAD + SE w/ Random (ours), RAD + SE w/ Contrastive, and RAD + SE w/ Inverse dynamics in Figure 7. We consider each backward pass to require twice as many FLOPs as a forward pass, as done in \url{https://openai.com/blog/ai-and-compute/}. Each weight requires one multiply-add operation in the forward pass. In the backward pass, it requires two multiply-add operations: at layer $i$, the gradient of the loss with respect to the weight at layer $i$ and with respect to the output of layer ($i-1$) need to be computed. The latter computation is necessary for subsequent gradient calculations for weights at layer ($i-1$). 

We use functions from \citet{huang2018condensenet} and \citet{jeong2019training} to obtain the number of operations per forward pass for all layers in the encoder (denoted $E$) and number of operations per forward pass for all MLP layers (denoted $M$). 

We assume that (1) the number of updates per iteration is 1, (2) the architecture of the encoder used for state entropy estimation is the same as the RL encoder used in RAD, and (3) the FLOPs required for computations (e.g., finding the $k$-NNs in representation space) that are not forward and backward passes through neural network layers is negligible.\footnote{Letting the dimension of $y$ be $d$, batch size $m$, computing distances between $y_{i}$ and all entries $y \in \mathcal{B}$ over $250000$ training steps requires $\sum^{250000}_{n=1000}(d(2m + 2\cdot\min(n, |\mathcal{B}|) + 3m\cdot\min(n, |\mathcal{B}|)) + 2m\cdot\min(n, |\mathcal{B}|))$, which is 1.569e+15 FLOPs in our case of $m = 512$, $|\mathcal{B}|=100000$, and $d=50$.}

We denote the number of forward passes per training update $F$, the number of backward passes per training update $B$, and the batch size $b$ (in our experiments $b = 512$). Then, the number of FLOPs per iteration of RAD is:
\begin{center}
    $bF(E + M) + 2bB(E + M) + (E + M),$
\end{center}
where the last term is for the single forward pass required to compute the policy action.

Specifically, RAD + SE w/ Random only requires $E$ extra FLOPs per iteration to store the fixed representation from the random encoder in the replay buffer for future $k$-NN calculations. In comparison, RAD + SE w/ Contrastive requires $4bE$ extra FLOPs per iteration, and RAD + SE w/ Inverse dynamics requires more than $6bE$ extra FLOPs. 

\clearpage

\section{Comparison to State Entropy Maximization with Contrastive Encoder for MiniGrid Pre-training}
\label{appendix:minigrid_atc_results}
\begin{figure*} [htb] \centering
\includegraphics[width=0.315\textwidth]{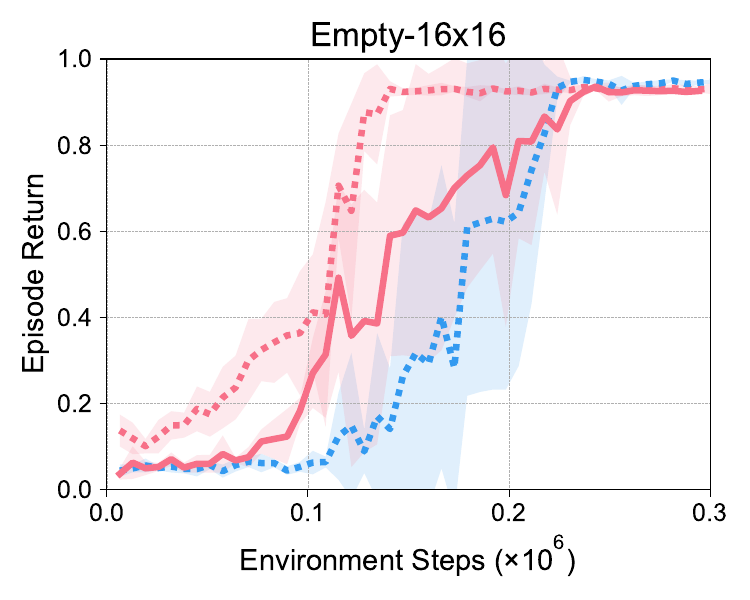}
\includegraphics[width=0.315\textwidth]{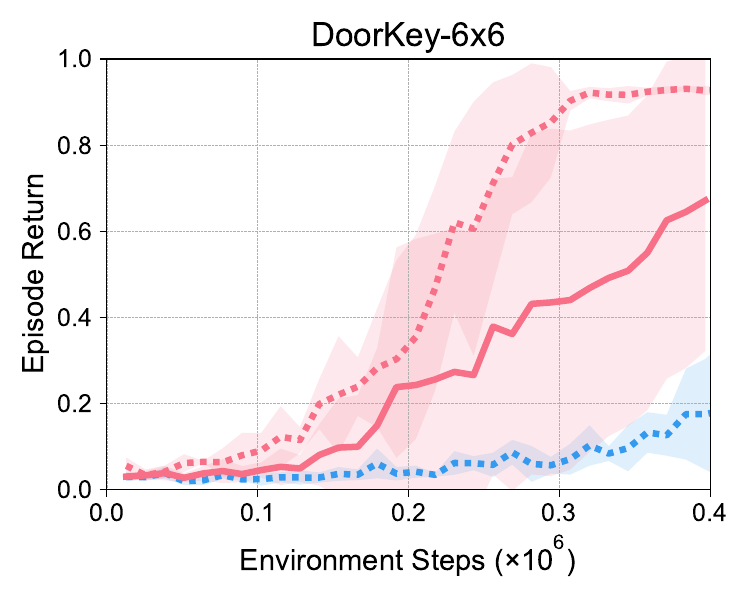}
\includegraphics[width=0.315\textwidth]{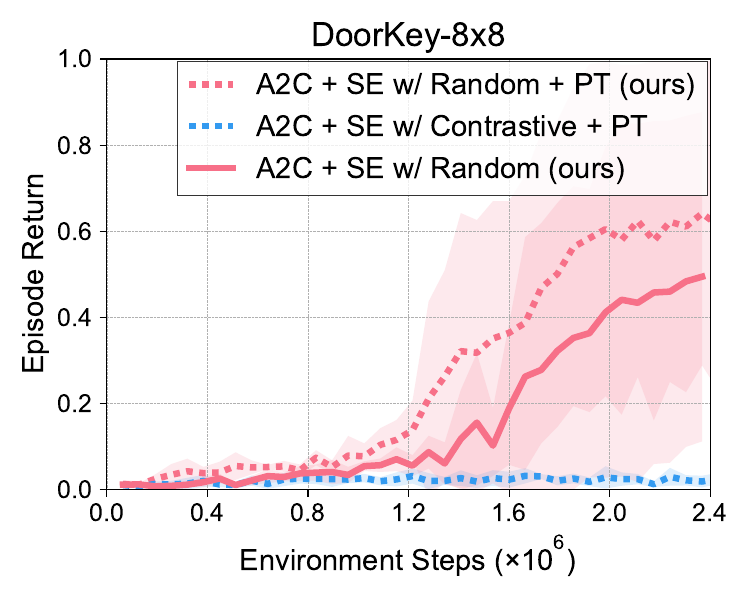}
\caption{Performance on navigation tasks from MiniGrid. We find that pre-training using state entropy (SE) with a random encoder (ours) outperforms pre-training using state entropy with a contrastive encoder \citep{liu2021behavior}, using random shift as the data augmentation. The solid (or dotted) line and shaded regions represent the mean and standard deviation, respectively, across five runs.}
\label{fig:minigrid_with_apt}
\end{figure*}

We show a comparison to state entropy maximization with contrastive learning (i.e., A2C + SE w/ Contrastive + PT) in Figure \ref{fig:minigrid_with_apt}. In the results shown, we do not use state entropy intrinsic reward during the fine-tuning phase for A2C + SE w/ Contrastive + PT, following the setup of \citet{liu2021behavior}, but found that using intrinsic reward during fine-tuning results in very similar performance. As discussed in Section 4.2, we observe that the contrastive encoder does not work well for state entropy estimation, as contrastive learning depends on data augmentation specific to images (e.g., random shift, random crop, and color jitter), which are not compatible with the compact embeddings used as inputs for MiniGrid. We re-mark that RE3 eliminates the need for carefully chosen data augmentations by employing a random encoder. 

\section{RE3 with On-policy Reinforcement Learning}
\label{appendix:on_policy_alg}
We provide the full procedure for RE3 with on-policy RL in Algorithm~\ref{alg:on_policy}.

\begin{algorithm}[htb]
\caption{RE3: On-policy RL version} \label{alg:on_policy}
\begin{algorithmic}[1]
\STATE Initialize parameters of random encoder $\theta$, policy $\phi$
\STATE Initialize replay buffer $\mathcal{B} \leftarrow \emptyset$, step counter $t \leftarrow 0$
\REPEAT
\STATE {{\textsc{// Collect transitions}}}
\STATE $t_{\tt{start}} \leftarrow t$
\REPEAT
\STATE Collect a transition $\tau_{t} = (s_{t}, a_{t}, s_{t+1}, r^{\tt{e}}_{t})$ from the interaction with the environment using policy $\pi_{\phi}$
\STATE Get a fixed representation $y_{t} = f_{\theta}(s_{t})$
\STATE $\mathcal{B} \leftarrow \mathcal{B} \cup \{(\tau_{t}, y_{t})\}$
\STATE $t \leftarrow t + 1$
\UNTIL{terminal $s_{t}$ or $t - t_{\tt{start}} = t_{\tt{max}}$}
\STATE {{\textsc{// Compute intrinsic reward}}}
\FOR{$j=t-1$ {\bfseries to} $t_{\tt{start}}$}
\STATE Compute the distance $||y_{j} - y||_{2}$ for all representations $y \in \mathcal{B}$ and find the $k$-nearest neighbor $y_{j}^{\text{$k$-NN}}$
\STATE Compute $r_{j}^{\tt{i}} \leftarrow \log(||y_{j} - y_{j}^{\text{$k$-NN}}||_{2} + 1)$
\STATE Let $r^{\tt{total}}_{j} \leftarrow r_{j}^{\tt{e}} + \beta \cdot r_{j}^{\tt{i}}$
\ENDFOR
\STATE {{\textsc{// Update policy}}}
\STATE Update $\phi$ with transitions $\{(s_{j}, a_{j}, s_{j+1}, r_{j}^{\tt{total}})\}_{j=t_{\tt{start}}}^{t-1}$
\UNTIL{$t > t_{\tt{max}}$}
\end{algorithmic}
\end{algorithm}

\end{document}